\documentclass{article}

\usepackage[eandd, final]{neurips_2026}

\usepackage[utf8]{inputenc} 
\usepackage[T1]{fontenc}    
\usepackage{hyperref}       
\usepackage{url}            
\usepackage{booktabs}       
\usepackage{amsfonts}       
\usepackage{nicefrac}       
\usepackage{microtype}      
\usepackage{xcolor}         
\usepackage{multirow}
\usepackage{graphicx}
\usepackage{pifont}
\usepackage{amsmath}
\usepackage{amssymb}
\usepackage{tikz}
\usetikzlibrary{calc}
\usepackage{subcaption}
\usepackage{makecell} 
\usepackage[colorinlistoftodos,prependcaption,textsize=tiny]{todonotes}
\usepackage{natbib}
\usepackage{wrapfig}
\usepackage{amssymb}
\usepackage{pdflscape}
\usepackage[super]{nth}
\usepackage{cleveref}
\usepackage{algorithm}
\usepackage{algpseudocode}
\usepackage{algorithmicx}
\usepackage[table]{xcolor}

\usepackage{amsmath}
\usepackage{booktabs}
\usepackage{makecell}
\usepackage{xcolor}

\newsavebox{\captiontable}

\newcommand{\dltp}[1]{\,{\scriptsize\color{green}$(+#1)$}}   
\newcommand{\dltm}[1]{\,{\scriptsize\color{red}$(-#1)$}}     
\newcommand{\dltz}{\,{\scriptsize$(\pm 0.00)$}}              

\hypersetup{
    colorlinks=true,
    citecolor=[rgb]{0.0,0.0,0.55},
    linkcolor=[rgb]{0.0,0.0,0.55},
    urlcolor=[rgb]{0.0,0.0,0.55}
}

\definecolor{green}{HTML}{10B981}
\definecolor{purple}{HTML}{3B82F6}
\definecolor{red}{HTML}{F43F5E}

\title{Beyond Self-Play and Scale: A \underline{Behavior Bench}mark
\\ for Generalization in Autonomous Driving}

\author{%
  Aron Distelzweig\,\textsuperscript{\normalfont 1, 2}
  \enskip
  Faris Janjo\v{s}\,\textsuperscript{\normalfont 2}
  \enskip
  Andreas Look\,\textsuperscript{\normalfont 3}
  \enskip
  Anna Rothenhäusler\,\textsuperscript{\normalfont 1}
  \enskip
  Daniel Jost\,\textsuperscript{\normalfont 1} 
  \\
  \textbf{Oliver Scheel}\,\textsuperscript{\normalfont 2}
  \enskip
  \textbf{Raghu Rajan}\,\textsuperscript{\normalfont 1}
  \enskip
  \textbf{Daphne Cornelisse}\,\textsuperscript{\normalfont 4} 
  \enskip
  \textbf{Eugene Vinitsky}\,\textsuperscript{\normalfont 4}
  \enskip
  \textbf{Joschka Boedecker}\,\textsuperscript{\normalfont 1} \\
  \textsuperscript{\normalfont 1} University of Freiburg  \enskip
  \textsuperscript{\normalfont 2} Bosch Center for Artificial Intelligence \\ 
  \textsuperscript{\normalfont 3} Coburg University of Applied Sciences \enskip
  \textsuperscript{\normalfont 4} New York University \\
  }

\begin{document}

\maketitle

\begin{abstract}
Recent Autonomous Driving (AD) works such as GigaFlow and PufferDrive have unlocked Reinforcement Learning (RL) at scale as a training strategy for driving policies. Yet such policies remain disconnected from established benchmarks, leaving the performance of large-scale RL for driving on standardized evaluations unknown.
We present \underline{\textbf{BehaviorBench}} -- a comprehensive test suite that closes this gap along three axes: \textcolor{green}{\textbf{Evaluation}}, \textcolor{purple}{\textbf{Complexity}}, and \textcolor{red}{\textbf{Behavior Diversity}}. In terms of \textcolor{green}{\textbf{Evaluation}}, we provide an interface connecting PufferDrive to nuPlan, which, for the first time, enables policies trained via RL at scale to be evaluated on an established planning benchmark for autonomous driving. Complementarily, we offer an evaluation framework that allows planners to be benchmarked directly inside the PufferDrive simulation, at a fraction of the time.
Regarding \textcolor{purple}{\textbf{Complexity}}, we observe that today's standardized benchmarks are so simple that near-perfect scores are achievable by straight lane following with collision checking. We extract a meaningful, interaction-rich split from the Waymo Open Motion Dataset (WOMD) on which strong performance is impossible without multi-agent reasoning. 
Lastly, we address \textcolor{red}{\textbf{Behavior Diversity}}. Existing benchmarks commonly evaluate planners against a single rule-based traffic model, the Intelligent Driver Model (IDM). We provide a diverse suite of interactive traffic agents to stress-test policies under heterogeneous behaviors, beyond just using IDM. Overall, our benchmarking analysis uncovers the following insight: despite learning interactive behaviors in an emergent manner, policies trained via pure self-play under standard reward functions overfit to their training opponents and fail to generalize to other traffic agent behaviors.
Building on this observation, we propose a hybrid planner that combines a PPO policy with a rule-based planner, providing a baseline for our new benchmark.
Code is available at \url{https://github.com/boschresearch/behavior-bench/}.
\end{abstract}
\begin{figure}[t]
    \centering
    \begin{subfigure}[t]{0.25\textwidth}
        \centering
        \begin{tikzpicture}
            \node[anchor=south west, inner sep=0] (img) at (0,0)
                {\raisebox{8pt}{\includegraphics[width=\textwidth]{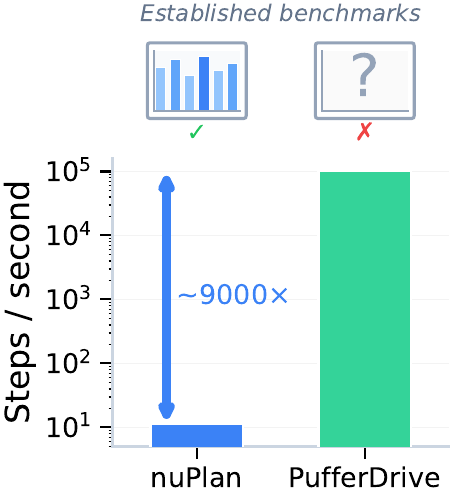}}};
            \begin{scope}[x={(img.south east)},y={(img.north west)}]
            \end{scope}
        \end{tikzpicture}
        \caption{\textcolor{green}{\textbf{Evaluation}}}
        \label{fig:throughput}
    \end{subfigure}
    \hfill
    \begin{subfigure}[t]{0.42\textwidth}
        \centering
        \begin{tikzpicture}
            \node[anchor=south west, inner sep=0] (img) at (0,0)
                {\raisebox{0pt}{\includegraphics[width=\textwidth]{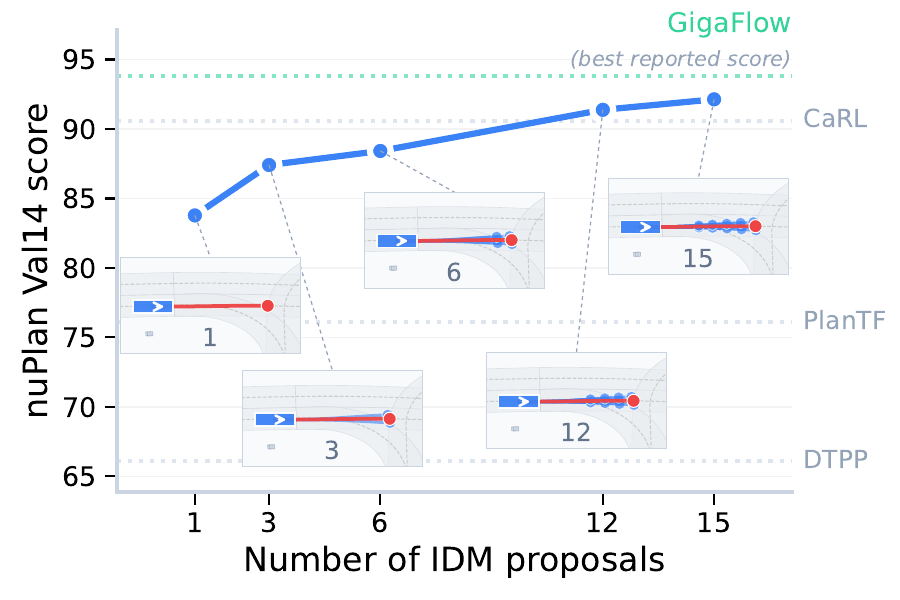}}};
            \begin{scope}[x={(img.south east)},y={(img.north west)}]
            \end{scope}
        \end{tikzpicture}
        \caption{\textcolor{purple}{\textbf{Benchmark complexity}}}
        \label{fig:complexity}
    \end{subfigure}
    \hfill
    \begin{subfigure}[t]{0.30\textwidth}
        \centering
        {\raisebox{12pt}{\includegraphics[width=\textwidth]{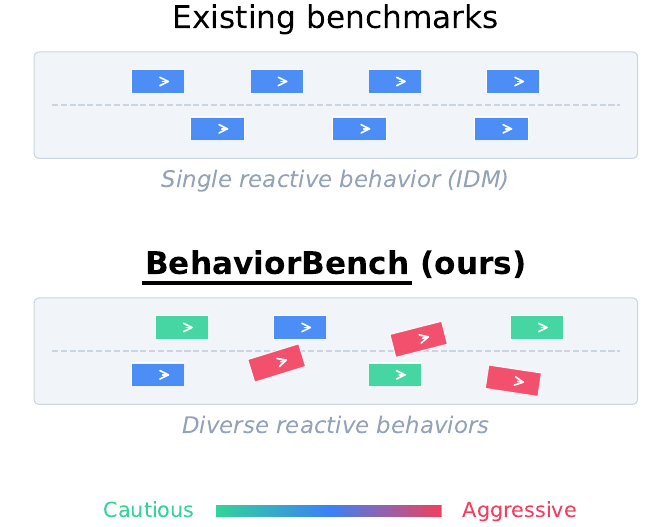}}}
        \caption{\textcolor{red}{\textbf{Traffic behavior diversity}}}
        \label{fig:diversity}
    \end{subfigure}
    \caption{Three limitations of current planner benchmarks. (a) \textcolor{green}{\textbf{Evaluation.}} PufferDrive achieves roughly 9000× higher step rates than nuPlan, making large-scale RL training practical for the first time. Without a matching evaluation framework, however, policies trained this way remain isolated from established benchmarks. (b) \textcolor{purple}{\textbf{Benchmark complexity.}} On the widely used nuPlan Val14 benchmark, a simple rule-based planner combining lane following with collision checking is already sufficient to achieve competitive scores: with only 15 IDM proposals, it clearly surpasses state-of-the-art learned planners such as CaRL~\citep{Jaeger2025CoRL}, PlanTF~\citep{cheng2024plantf} and DTPP~\citep{huang2024dtpp}. This shows that the benchmark barely measures interactive or negotiation capabilities and systematically overestimates true planner performance. (c) \textcolor{red}{\textbf{Traffic behavior diversity.}} Existing benchmarks evaluate almost exclusively against a single reactive traffic model (IDM), whose agents are predictable, cooperative, and uniform (top). Our framework instead introduces a heterogeneous set of traffic agents (bottom), enabling, for the first time, a cross-agent evaluation that distinguishes genuine generalization from overfitting to a single behavioral regime.}
    \label{fig:overview}
    \vspace{-0.6cm}
\end{figure}

\section{Introduction}
The ability to train policies at scale is unlocking new capabilities in Autonomous Driving (AD)~\citep{cusumanotowner2025gigaflow}. Reinforcement Learning (RL), in particular, promises to discover policies that go beyond the limitations of both rule-based and imitation-learned planners by optimizing directly for driving objectives. However, it requires millions of environment interactions, far exceeding what current simulators can deliver within a reasonable amount of time. 
Existing simulation platforms each address a different aspect of AD research, yet none enable large-scale RL. nuPlan~\citep{nuplan} targets Imitation Learning (IL) and planner evaluation, CARLA~\citep{Dosovitskiy17CARLA} provides photorealistic sensor rendering, and Waymax~\citep{waymax} accelerates multi-agent simulation but still falls short of the throughput required to train policies at scale.

A notable exception is PufferDrive~\citep{cornelisse2025pufferdrive}, which was recently introduced to break through this barrier. By combining a batched simulation architecture with a highly optimized engine written in C~\citep{suarez2024pufferlib}, it achieves orders-of-magnitude higher throughput than prior simulators (Fig.~\ref{fig:throughput}), making large-scale training of driving policies practical for the first time. Yet speed alone is not enough. Without standardized benchmarks, reproducible evaluation protocols, and compatibility with established evaluation suites, policies trained in PufferDrive exist in isolation -- they cannot be compared against the state of the art, and their true capabilities remain unknown.

A structured comparison of these simulation frameworks, highlighting the trade-offs between simulator capabilities and computational throughput, is provided in Appendix~\ref{sec:related_work}, to which we refer the reader for an exhaustive discussion of relevant simulation environments.

\textbf{In this work, we close this gap by introducing the first standardized benchmark and evaluation framework for PufferDrive.} Our framework supports \textcolor{green}{\textbf{evaluation}} directly within PufferDrive and provides an interface for evaluating PufferDrive-trained policies on established benchmarks such as nuPlan~\citep{nuplan} and interPlan~\citep{hallgarten2024interplan}, allowing direct comparison with prior work.
Crucially, however, we go beyond mere compatibility and address \textcolor{purple}{\textbf{benchmark complexity}}. We argue that the existing benchmarks themselves are insufficient. The most widely-used nuPlan Val14 benchmark~\citep{Dauner2023CORL}, can be largely solved by lane-following combined with simple collision-checking (Fig.~\ref{fig:complexity}). Prior work has demonstrated that even a small set of rule-based proposals suffices to achieve strong performance~\citep{Dauner2023CORL, distelzweig2025spdm}. This reveals a fundamental problem -- planners can obtain high scores without reasoning about interactions, anticipating the behavior of surrounding agents, or negotiating in congested traffic situations.

This problem is compounded by a second, equally critical limitation: \textcolor{red}{\textbf{traffic behavior diversity}} (Fig.~\ref{fig:diversity}).   Most existing benchmarks evaluate planners against a single reactive traffic model, the rule-based Intelligent Driver Model (IDM)~\citep{Treiber2000idm}. IDM agents follow their lane, react through simple car-following dynamics, and rarely exhibit the diverse, sometimes adversarial behaviors encountered in real traffic. As a consequence, benchmark scenarios become artificially easy -- surrounding traffic is predictable and cooperative, and a planner may appear robust simply because it has adapted to the narrow assumptions of this one traffic model. This issue is particularly pronounced for RL-trained planners, which are not only evaluated but often trained against the same traffic model. Throughout training, the policy continuously interacts with this single behavioral model and can readily overfit to its regularities, learning to exploit IDM-specific dynamics rather than acquiring strategies that transfer to other behaviors. Evaluation under the same behavioral regime therefore fundamentally cannot completely assess generalization.

interPlan~\citep{hallgarten2024interplan} attempted to address the first issue by augmenting nuPlan with handcrafted out-of-distribution scenarios. Although a step in the right direction, these scenarios are limited in both scale and diversity. They cover only a small number of maps and traffic configurations and still rely on the same underlying IDM traffic behavior.
We address both shortcomings simultaneously. First, we construct a substantially larger and more challenging benchmark based on the Waymo Open Motion Dataset (WOMD)~\citep{waymo_dataset}. Rather than curating a handful of edge cases, we systematically extract a collection of 1{,}000 highly interactive situations, including dense traffic, merges, blocked lanes, unprotected turns, and other complex multi-agent interactions, in which simple lane following is no longer sufficient. To succeed, a planner must actively reason about surrounding agents and adapt its behavior.

Second, we move beyond the single-traffic-model paradigm by introducing a diverse set of traffic agents with distinct driving styles and interaction patterns. Some agents yield cautiously while others drive aggressively and force interactions. Evaluating planners across this spectrum of traffic dynamics allows us, for the first time, to measure whether a policy genuinely generalizes or merely overfits to a particular behavioral regime.
We validate our framework through extensive experiments and find that planners achieving strong results on existing benchmarks frequently fail in our interaction-heavy scenarios and under alternative traffic agents. These results demonstrate that current benchmarks substantially overestimate planner capabilities and are blind to a critical dimension of robustness.

Our contribution is a comprehensive benchmarking framework for planner evaluation in PufferDrive, consisting of the following components:
\begin{itemize}
    \item The first comprehensive \textcolor{green}{\textbf{evaluation interface}} for planners in PufferDrive, including a dedicated PufferDrive--nuPlan interface that, for the first time, enables \emph{at-scale trained} RL policies to be evaluated inside the established nuPlan simulator, bridging the gap between at-scale RL training and standardized planner benchmarking.
    \item  \textcolor{purple}{\textbf{Two curated 1k scenario splits}} based on WOMD that systematically cover a broad spectrum of traffic situations, from routine driving to highly interactive scenarios.
    \item A \textcolor{red}{\textbf{diverse suite of traffic agents}} enabling comprehensive cross-agent evaluation and exposing planner behavior under heterogeneous and realistic interaction patterns.
\end{itemize}
Using this framework, we derive the following findings and methodologies:
\begin{itemize}
    \item Empirical evidence that self-play RL agents trained at-scale fail to generalize to unseen traffic agents.
    \item A hybrid planner that leverages the reliability of rule-based approaches and establishes a strong baseline on our benchmark.
\end{itemize}
\section{BehaviorBench}
\begin{figure}[ht]
    \includegraphics[width=\textwidth]{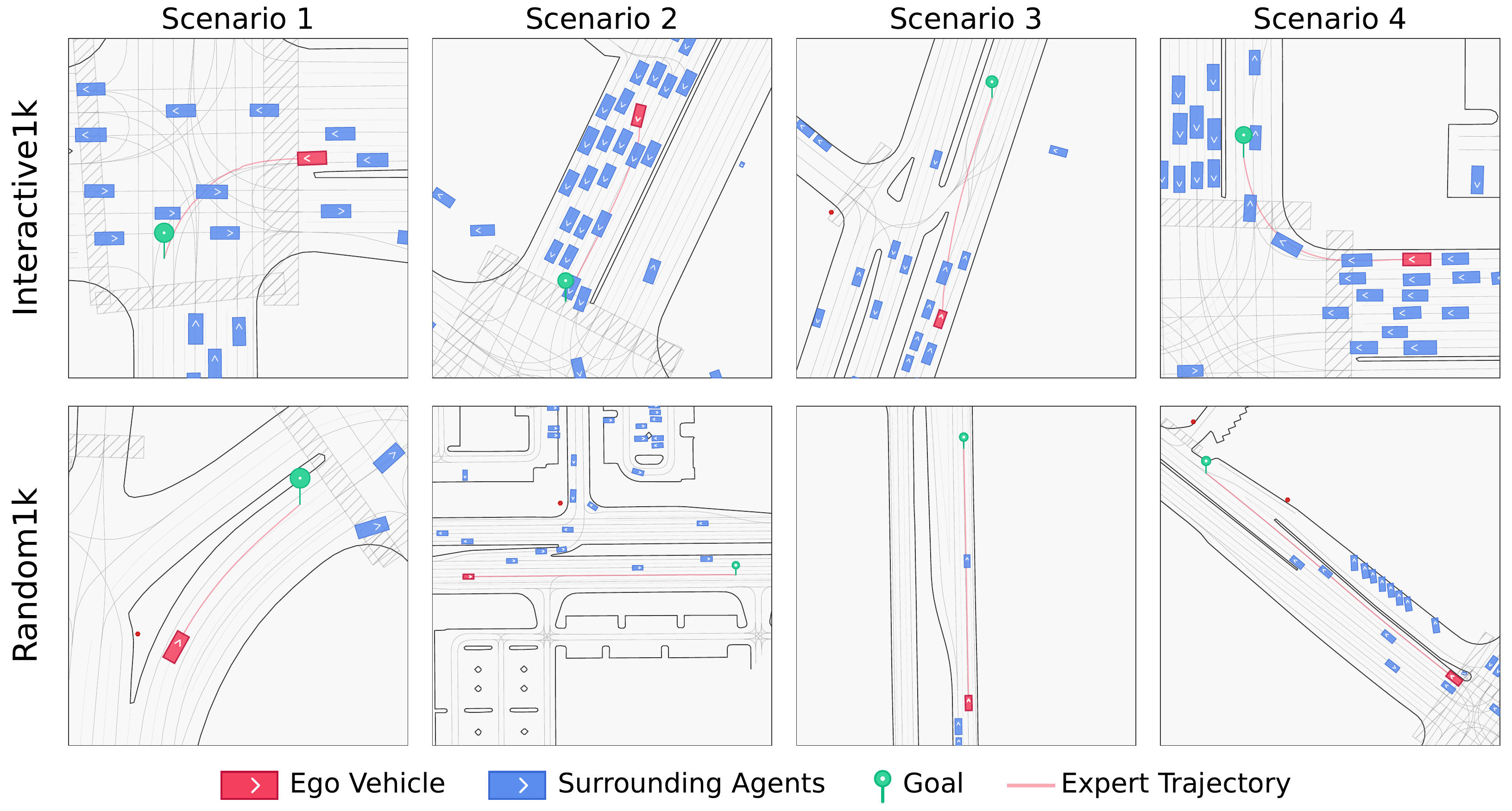}
    \caption{\textbf{Qualitative comparison of our Interactive1k and Random1k splits.}
Four representative scenarios from each split, showing the ego vehicle (red), surrounding agents (blue), the ego goal (green), and the expert trajectory (red). \emph{Interactive1k} (top row) is dominated by dense urban situations while \emph{Random1k} (bottom row) reflects simpler cases.}
    \vspace{-0.5cm}
    \label{fig:qualitative_splits}
\end{figure}

We introduce \underline{\textbf{BehaviorBench}}, a benchmark built around three components. Sec.~\ref{subsec:framework} describes the \textcolor{green}{\textbf{evaluation interface}} and its interface to nuPlan, which for the first time enables at-scale RL-trained policies to be evaluated on an established planning benchmark. Sec.~\ref{subsec:splits} introduces \textcolor{purple}{\textbf{two curated 1k scenario splits}} that jointly cover routine and highly interactive driving situations. Sec.~\ref{subsec:agents} presents our \textcolor{red}{\textbf{diverse suite of traffic agents}} that measures planner performance under heterogeneous and realistic interaction patterns. Finally, Sec.~\ref{subsec:pdm_ppo} introduces our hybrid planner.

\subsection{\textcolor{green}{Evaluation Interface}}
\label{subsec:framework}
\paragraph{Simulation environment.} PufferDrive is a high-throughput driving simulator built on a vectorized C backend, originally designed purely as a training environment without support for evaluating trained policies against independent traffic models. Our framework closes this gap. Each scenario contains up to 128 agents. The simulator exposes ego-centric observations comprising the ego state ($\mathbb{R}^{7}$), the 31 nearest surrounding agents ($\mathbb{R}^{31 \times 8}$), and 128 visible road segments ($\mathbb{R}^{128 \times 7}$). Actions are available either as a discrete combination of acceleration and steering or as a discrete combination of longitudinal and lateral jerk, integrated through a kinematic bicycle model at $\Delta t = 0.1$\,s. Vulnerable road users (pedestrians and cyclists) are treated as non-controllable participants that replay their logged expert trajectories. Agents are removed upon reaching a terminal condition: goal completion, leaving the drivable area (off-road), or collision.

\paragraph{Evaluation protocol.} In our proposed evaluation framework, for every scenario, a single agent is designated as the ego and controlled by the planner, while the remaining agents are controlled by a traffic agent model. An episode terminates as soon as the ego reaches its goal, collides with another agent, or leaves the drivable area.
Our framework implements nuPlan's \texttt{AbstractPlanner} interface, allowing any PufferDrive planner to be evaluated under nuPlan's simulation and its official metrics suite. We additionally support interPlan evaluation. This compatibility enables direct comparison with prior work on both benchmarks.

\paragraph{Benchmark Score.}
To quantify overall planner performance, we combine three task-critical outcome indicators and three behavior scores into a single per-scenario value in $[0, 1]$. The outcome terms are combined multiplicatively, while the behavior terms are aggregated as a weighted sum
\begin{equation}
    \text{Score}_{\text{scenario}} \;=\;
    \underbrace{\bigl(1 - \mathbb{I}[\text{AF-Col}]\bigr)\,\bigl(1 - \mathbb{I}[\text{Off-Road}]\bigr)\,\mathbb{I}[\text{Goal}]}_{\text{outcome indicator} \,\in\, \{0, 1\}}
    \;\cdot\;
    \underbrace{\bigl(w_{\text{cmf}}\,S_{\text{cmf}} + w_{\text{align}}\,S_{\text{align}} + w_{\text{ctr}}\,S_{\text{ctr}}\bigr)}_{\text{behavior score} \,\in\, [0, 1]},
    \label{eq:benchmark-score}
\end{equation}
with weights $w_{\text{cmf}} = 0.2$,  $w_{\text{align}}=0.5$, and $w_{\text{ctr}} = 0.3$. Indicators $\mathbb{I}[\cdot] \in \{0, 1\}$ mark whether the ego caused an At-Fault collision (AF-Col), left the drivable area (Off-Road), or reached the goal during the scenario. The multiplicative formulation enforces that all three outcome criteria must be satisfied as failure in any one collapses the score to zero. Within successful scenarios, the weighted sum rewards comfort ($w_{\text{cmf}}$), lane-alignment ($w_{\text{align}}$) and well-centered ($w_{\text{ctr}}$) driving.
In summary, our score jointly captures whether the drive succeeded and how well it was executed.
We report the mean across $N$ scenarios as: $\text{Score} = \frac{1}{N} \sum_{n=1}^{N} \text{Score}_{\text{scenario}}^{(n)}$. Detailed explanation of subscores are in Appendix~\ref{sec:eval-metrics}.

\subsection{\textcolor{purple}{Evaluation Splits}}
\label{subsec:splits}
Following existing data-driven simulators~\citep{nuplan, waymax}, we initialize scenarios from real-world driving logs. We leverage the WOMD, traditionally used for \textit{open-loop} evaluation, as a source of initial conditions for the \textit{closed-loop} evaluation of driving policies, and filter their interaction-rich recordings to construct challenging scenarios. This is motivated by the observation that widely used planning benchmarks such as nuPlan Val14~\citep{Dauner2023CORL} are comparatively easy: as shown in Fig.~\ref{fig:complexity}, even a small number of IDM proposals is sufficient to essentially solve them, leaving little headroom to distinguish competing planners. We curate an \textit{Interactive} and a \textit{Random} split of 1{,}000 scenarios each from the WOMD validation set. Qualitative examples from both splits are shown in Fig.~\ref{fig:qualitative_splits}, with additional examples in Fig.~\ref{fig:app_interactive1k} and Fig.~\ref{fig:app_random1k}.

We define an \emph{interactivity score} $S_{\text{int}} \in [0,1]$ to select the most challenging scenarios. The score is a weighted sum of normalized components, scaled by a lane-change multiplier
\begin{equation}
S_{\text{int}} = \lambda_{\text{lane}} \sum_{i} w_i \cdot \min\left(\frac{c_i}{\tau_i}, 1\right),
\end{equation}
where each component $c_i$ captures a distinct aspect of scenario difficulty and is clipped to $[0,1]$ via a normalization threshold $\tau_i$. The lane-change multiplier $\lambda_{\text{lane}}$ take the value $1.0$ if reaching the goal requires a lane change or turn, and $0.5$ otherwise. The components $c_i$ are: number of unique agents whose ground-truth trajectories intersect with the ego's ground-truth trajectory, ego acceleration and steering changes, number of simulation steps where the ego's time-to-collision with an agent falls below 3\,s, number of agents within a 40\,m radius and distance to the ego's goal. We provide more details in Sec.~\ref{sec:i1k}. We select the 1{,}000 scenarios with the highest $S_{\text{int}}$ to form the \emph{Interactive1k split}. The weights $w_i$ and normalization thresholds $\tau_i$ can be found in Tab.~\ref{tab:score_components}.
To complement the Interactive1k split and retain coverage of the broader driving distribution, we additionally construct a \textit{Random1k} split by uniformly sampling 1{,}000 scenarios from the WOMD validation set.

\subsection{\textcolor{red}{Traffic Agents}}
\label{subsec:agents}
Unlike most existing benchmarks, which are limited to a single reactive traffic model, we provide a heterogeneous set of reactive agents. Each agent model is described in detail below.

\paragraph{Intelligent Driver Model (IDM).}
A rule-based model~\citep{Treiber2000idm} that computes acceleration based on the current velocity, desired speed, and distance to the leading vehicle as
\begin{equation}
\frac{dv}{dt} = a\left[1 - \left(\frac{v}{v_0}\right)^{\delta} - \left(\frac{s^{*}}{s}\right)^{2}\right] .
\end{equation}
Intuitively, the policy accelerates with $a$ unless the velocity is already close to the target speed $v_0$ or the leading vehicle is within the safety margin $s^{*}$. Lateral guidance follows lane centerlines. Our choice of parameters is listed in Tab.~\ref{tab:idm-params}. 

\paragraph{Conditioned PPO.}
A decentralized policy $\pi_\theta^{c}$ trained with Proximal Policy Optimization (PPO)~\citep{schulman2017ppo} and self-play, i.e. controlling all agents in a scene simultaneously via parameter sharing. Network architecture details are provided in Sec.~\ref{sec:architecture}.
Following GigaFlow~\citep{cusumanotowner2025gigaflow}, we train the policy with a complex reward and \emph{reward conditioning}, exposing the reward weights as an input to the network rather than optimizing against a fixed reward. For every episode, each agent samples a ten-dimensional weight vector $c \sim p(c)$, and the PPO reward is assembled from those weights (reward function in Appendix~\ref{sec:reward-baselines}, training distribution in Appendix~\ref{sec:reward-conditioning}). Because $c$ is appended to the observation, the policy learns to map each coefficient setting to the behavior optimal \emph{under that reward}, yielding a continuous family of driving styles within the same parameters $\theta$. At evaluation time, fixing $c$ to any target point within the sampling range deterministically selects the associated behavior. We exploit this to construct three traffic agents differing only in $c$, namely \textcolor{red}{\emph{Aggressive}}, \textcolor{purple}{\emph{Normal}}, and \textcolor{green}{\emph{Cautious}}, plus a round-robin mixture so a single scene can contain all three styles. Coefficient values are listed in Tab.~\ref{tab:traffic-profiles}. These three profiles are merely representative samples of a continuum, since any point in the sampling range elicits a corresponding behavior, and the same trained policy can in principle instantiate an unbounded family of traffic agents at no additional training cost.

\paragraph{PPO.}
In addition to the reward conditioned network, we train a network without the coefficient vector as input and with a simple reward, we provide more details on the reward in Appendix~\ref{sec:reward-baselines}.

\paragraph{SMART.}
An autoregressive decoder-only motion prediction model~\citep{wu2024smart} that ranked \nth{1} on the Waymo Open Sim Agents Challenge 2024. We use the 1-million-parameter variant, trained via behavior cloning on the WOMD training set. SMART jointly predicts trajectories for all agents in a scene by tokenizing motion into a learned codebook of primitives, where each token represents 0.5\,s of movement. At each rollout step, the model predicts the next token for every agent simultaneously, the corresponding 0.5\,s of motion is then executed in simulation before the next prediction step.

\paragraph{Expert (Log-Replay).}
Each agent replays its recorded ground-truth trajectory from WOMD. Because replay is open-loop, traffic does not react to the ego vehicle's actions.

\subsection{PDM+PPO: Bridging Rule-based and RL Planners}
\label{subsec:pdm_ppo}
We introduce a hybrid planner that combines a learned PPO policy with a rule-based component. For the latter, we adopt the Predictive Driver Model (PDM-Closed)~\citep{Dauner2023CORL}, winner of the nuPlan Challenge 2023. PDM predicts surrounding agents under a constant-velocity assumption over a $4$\,s horizon, generates $15$ IDM proposals with varying target velocities, scores each by traffic-rule compliance, progress, and comfort, and executes the best-scoring trajectory. By design, PDM cannot perform lane changes and therefore tends to behave overly conservatively.

Our hybrid planner couples PDM with the PPO policy from Sec.~\ref{subsec:agents}. At each step, IDM proposals are generated and scored as in PDM, and if all receive a score of zero (every candidate either causes a collision, leaves the drivable area, or fails to make progress), control is delegated to PPO. The PPO policy is trained with a simple reward covering only collision avoidance, off-road avoidance, and goal completion, which requires almost no tuning but captures only high-level driving, ignoring behavioral aspects such as comfort and lane alignment, and tends to overfit to the training opponents.

To mitigate both issues, rather than executing the PPO action directly, we select its top-$K$ actions and roll each out for $1$\,s under constant-velocity predictions of the surrounding agents, scoring the resulting trajectories with Eq.~\ref{eq:benchmark-score}. This switching mechanism preserves PDM's safety in routine situations while leveraging the more assertive PPO policy when PDM cannot identify a valid maneuver. Details are provided in Appendix~\ref{sec:app_pdm_ppo}.
\section{Experimental Results}
\begin{table}[t]
\centering
\small
\setlength{\tabcolsep}{12pt}
\renewcommand{\arraystretch}{1.15}
\resizebox{\textwidth}{!}{%
\begin{tabular}{@{}l cccc cccc@{}}
\toprule
& \multicolumn{8}{c}{\textbf{Traffic Agents}} \\
\cmidrule(lr){2-9}
\textbf{Planner}
& IDM & PPO & SMART & Expert
& \makecell{Aggressive \\ \textcolor{red}{$\pi_\theta^{\text{aggr}}$}}
& \makecell{Normal \\ \textcolor{purple}{$\pi_\theta^{\text{norm}}$}}
& \makecell{Cautious \\ \textcolor{green}{$\pi_\theta^{\text{caut}}$}}
& \makecell{Mix \\ $\{
{\color{red}\pi_\theta^{\text{aggr}}},
{\color{purple}\pi_\theta^{\text{norm}}},
{\color{green}\pi_\theta^{\text{caut}}}
\}$} \\
\midrule
\rowcolor{gray!12} \multicolumn{9}{c}{\textit{Interactive1k}} \\
\midrule
IDM            & 15.11 & 20.61 & 15.74 & 12.17 & 11.22 & 7.95 & 5.36 & 7.76 \\
PDM            & 24.11 & 29.23 & 30.44 & 20.68 & 20.76 & 14.02 & 8.42 & 12.69 \\
PPO            & \underline{45.08}\,{\scriptsize$\pm2.60$} & \underline{59.45}\,{\scriptsize$\pm1.43$} & \underline{47.23}\,{\scriptsize$\pm2.16$} & 37.07\,{\scriptsize$\pm3.13$} & \underline{46.47}\,{\scriptsize$\pm1.23$} & \underline{42.57}\,{\scriptsize$\pm1.87$} & \textbf{40.14}\,{\scriptsize$\pm2.62$} & \underline{42.57}\,{\scriptsize$\pm2.30$} \\
SMART          & 28.13 & 33.84 & 36.76 & \underline{38.29} & 23.68 & 18.77 & 9.96 & 14.92 \\
PDM+PPO (Ours) & \textbf{57.10}\,{\scriptsize$\pm3.44$} & \textbf{73.98}\,{\scriptsize$\pm2.10$} & \textbf{62.90}\,{\scriptsize$\pm3.95$} & \textbf{52.37}\,{\scriptsize$\pm5.58$} & \textbf{53.21}\,{\scriptsize$\pm3.24$} & \textbf{46.98}\,{\scriptsize$\pm2.99$} & \underline{39.10}\,{\scriptsize$\pm4.67$} & \textbf{44.88}\,{\scriptsize$\pm4.09$} \\
\addlinespace[3pt]
\midrule
\rowcolor{gray!12} \multicolumn{9}{c}{\textit{Random1k}} \\
\midrule
IDM            & 55.40 & 59.11 & 52.94 & 50.80 & 49.50 & 46.64 & 42.24 & 45.98 \\
PDM            & \underline{63.91} & \underline{67.10} & \underline{61.08} & \underline{56.87} & 55.40 & 53.58 & 47.66 & 51.60 \\
PPO            & 58.11\,{\scriptsize$\pm1.55$} & 64.64\,{\scriptsize$\pm0.80$} & 56.84\,{\scriptsize$\pm0.65$} & 53.43\,{\scriptsize$\pm1.39$} & \underline{58.00}\,{\scriptsize$\pm0.44$} & \underline{55.30}\,{\scriptsize$\pm0.68$} & \underline{52.31}\,{\scriptsize$\pm1.25$} & \underline{55.44}\,{\scriptsize$\pm0.93$} \\
SMART          & 48.13 & 48.46 & 46.55 & 50.07 & 41.65 & 38.93 & 32.88 & 36.60 \\
PDM+PPO (Ours) & \textbf{69.08}\,{\scriptsize$\pm1.32$} & \textbf{75.44}\,{\scriptsize$\pm1.32$} & \textbf{68.17}\,{\scriptsize$\pm1.26$} & \textbf{65.10}\,{\scriptsize$\pm0.99$} & \textbf{63.33}\,{\scriptsize$\pm0.72$} & \textbf{61.66}\,{\scriptsize$\pm0.89$} & \textbf{55.46}\,{\scriptsize$\pm1.30$} & \textbf{60.34}\,{\scriptsize$\pm1.19$} \\
\bottomrule
\end{tabular}%
}
\caption{\textbf{Benchmark scores across traffic agents on Interactive1k and Random1k.} Each column corresponds to a traffic agent type, and each row to a planner. Scores are computed according to Eq.~\ref{eq:benchmark-score} and scaled by 100. For PPO and PDM+PPO we report mean$\pm$std over three training runs with different seeds. The best result in each column is shown in \textbf{bold}, and the second-best result is \underline{underlined}.}
\vspace{-0.8cm}
\label{tab:planner_score}
\end{table}

\begin{figure}[t]
    \centering
    \includegraphics[width=\linewidth]{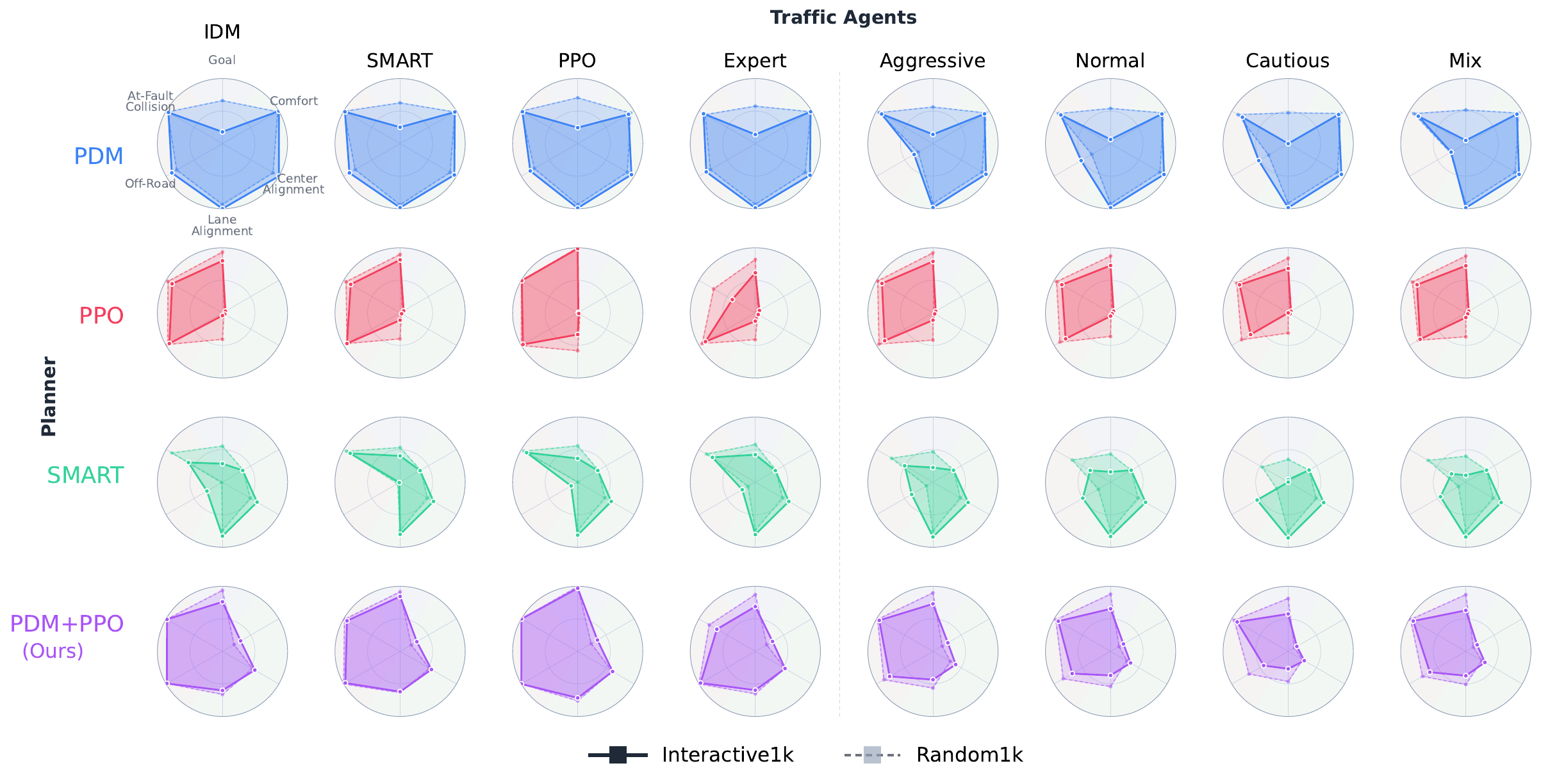}
    \caption{\textbf{Planner performance across traffic agents and benchmark splits.} Each radar plot shows the six core metrics, At-Fault Collision, Off-Road, Goal completion, Comfort, Center Alignment, and Lane Alignment, for a single (planner, traffic agent) combination. Rows correspond to the four planners and columns to the eight traffic agents. All axes are oriented such that the outer ring corresponds to the best observed value, with At-Fault Collision and Off-Road inverted. Each metric is independently min-max normalized across both splits and all (planner, traffic agent) cells, so axis ranges are directly comparable across the entire grid. Within each cell, the solid filled polygon shows performance on Interactive1k and the dashed polygon shows performance on Random1k. 
    }
    \vspace{-0.4cm}
    \label{fig:subscores_radar}
\end{figure}

\begin{table}[t]
\centering
\resizebox{1.0\columnwidth}{!}{
\begin{tabular}{lccccccccc}
\toprule
\multirow{2}{*}{Method}
& \multirow{2}{*}{Score $\uparrow$}
& \multicolumn{8}{c}{Subscores $\uparrow$} \\
\cmidrule(lr){3-10}
& & No-Col. & TTC & Making Progress & Progress along route ratio & Comfort & Drivable & Direction & Speed \\
\midrule
\multicolumn{10}{c}{\textit{nuPlan Val14}~\citep{Dauner2023CORL}} \\
\midrule
PDM~\citep{Dauner2023CORL} & 92.13 & 97.90 & \textbf{93.83} & 99.11 & 90.26 & \textbf{94.72} & 99.46 & \textbf{99.96} & \textbf{99.83} \\
DiffusionPlanner~\citep{zheng2025diffusionplanner} & 82.80  & -- & -- & -- & -- &-- & -- & -- & -- \\
PPO      & 67.24 & 91.46 & 62.52 & 99.11 & \textbf{92.26} & 39.18 & 99.19 & 92.13 & 93.76 \\
PDM+PPO (Ours)  & \textbf{92.49} & \textbf{98.39} & 93.11 & \textbf{99.37} & 91.12 & 91.95 & \textbf{99.55} & 99.51 & 99.79 \\

\midrule
\multicolumn{10}{c}{\textit{interPlan}~\citep{hallgarten2024interplan}} \\
\midrule
PDM~\citep{Dauner2023CORL}      & 41.68 & 91.25 & 90.00 & \textbf{81.25} & 44.91 & \textbf{93.75} & \textbf{100} & \textbf{100} & \textbf{100} \\
DiffusionPlanner~\citep{zheng2025diffusionplanner} & 25.76 & -- & -- & -- & -- &-- & -- & -- & -- \\
PPO      & 42.10 & 90.00 & 65.00 & \textbf{81.25} & \textbf{69.62} & 11.25 & 93.75 & 61.25 & 94.55 \\
PDM+PPO (Ours)  & \textbf{43.24} & \textbf{97.50} & \textbf{91.25} & 80.00 & 44.14 & 88.75 & \textbf{100} & 96.25 & \textbf{100} \\
\bottomrule
\end{tabular}}
\caption{\textbf{nuPlan and interPlan results.} Overall score and individual subscores on the nuPlan Val14 and interPlan reactive benchmarks. PPO is trained via self-play on nuPlan maps.}
\vspace{-1cm}
\label{tab:val14_interplan}
\end{table}

\begin{table}[t]
\centering
\small
\setlength{\tabcolsep}{4pt}
\renewcommand{\arraystretch}{1.1}
\resizebox{1\textwidth}{!}{
\begin{tabular}{l|cccccc|cccccc}
\toprule
\multirow{2}{*}{Method}
& \multicolumn{6}{c|}{Interactive1k}
& \multicolumn{6}{c}{Random1k} \\

\cmidrule(lr){2-7} \cmidrule(lr){8-13}

& \makecell{Realism Meta \\ Metrics$\uparrow$}
& \makecell{Kinematic \\ Metrics$\uparrow$}
& \makecell{Interactive \\ Metrics$\uparrow$}
& \makecell{Map-based \\ Metrics$\uparrow$}
& \makecell{minADE$\downarrow$}
& \makecell{ADE$\downarrow$}
& \makecell{Realism Meta \\ Metrics$\uparrow$}
& \makecell{Kinematic \\ Metrics$\uparrow$}
& \makecell{Interactive \\ Metrics$\uparrow$}
& \makecell{Map-based \\ Metrics$\uparrow$}
& \makecell{minADE$\downarrow$}
& \makecell{ADE$\downarrow$} \\

\midrule
Ground-truth
& 0.80 & 0.50 & 0.87 & 0.88 & 0 & 0
& 0.80 & 0.52 & 0.84 & 0.90 & 0 & 0 \\
\midrule
Random
& 0.41 & 0.05 & 0.53 & 0.46 & 32.53 & 33.95
& 0.41 & 0.04 & 0.52 & 0.47 & 27.96 & 29.39 \\

IDM
& 0.67 & 0.25 & 0.76 & 0.79 & 7.36 & 7.36
& 0.67 & 0.28 & 0.74 & 0.80 & 6.48 & 6.48 \\

PPO
& 0.64 & 0.18 & 0.72 & 0.79 & 3.69 & 6.47
& 0.64 & 0.20 & 0.72 & 0.80 & 4.21 & 6.62 \\

SMART
& \textbf{0.74} & \textbf{0.37} & \textbf{0.82} & \textbf{0.84} & \textbf{1.72} & \textbf{3.74}
& \textbf{0.74} &\textbf{ 0.40} & \textbf{0.80} & \textbf{0.85}& \textbf{1.56} & \textbf{3.63} \\

$\{
{\color{red}\pi_\theta^{\text{aggr}}},
{\color{purple}\pi_\theta^{\text{norm}}},
{\color{green}\pi_\theta^{\text{caut}}}
\}$
& 0.68 & 0.29 & 0.78 & 0.78 & 8.11 & 8.11
& 0.69 & 0.33 & 0.76 & 0.80 & 6.80 & 6.80 \\

\color{red}$\pi_\theta^{\text{aggr}}$
& 0.68 & 0.33 & 0.75 & 0.78 & 5.39 & 5.39
& 0.69 & 0.35 & 0.74 & 0.81 & 5.10 & 5.10 \\

\color{purple}$\pi_\theta^{\text{norm}}$
& 0.69 & 0.31 & 0.80 & 0.79 & 5.96 & 5.96
& 0.70 & 0.34 & 0.77 & 0.81 & 5.30 & 5.30 \\

\color{green}$\pi_\theta^{\text{caut}}$
& 0.68 & 0.26 & 0.80 & 0.78 & 14.13 & 14.13
& 0.68 & 0.30 & 0.77 & 0.80 & 10.92 & 10.92 \\

\bottomrule
\end{tabular}}
\caption{\textbf{WOSAC 2024 results on Interactive1k and Random1k.} Metrics include realism meta, kinematic, interactive, map-based scores and trajectory errors minADE and ADE.}
\vspace{-0.7cm}
\label{tab:realism_combined}
\end{table}

\paragraph{Benchmark results.}
Tab.~\ref{tab:planner_score} reports results on our scenario splits. Excluding our hybrid planner, the two splits induce contrasting rankings: PPO achieves the highest final score against every traffic agent on Interactive1k, while PDM shows stronger performance on Random1k compared to Interactive1k. PDM's rule-based design yields safe, conservative trajectories that suffice in routine driving but fail to resolve situations requiring active negotiation, as confirmed by its sharp goal-completion increase from Interactive1k to Random1k in Fig.~\ref{fig:subscores_radar}. PPO, trained with a goal-completion reward, favors assertive maneuvering that pays off in interactive scenes but offers no advantage in routine traffic. SMART sits in between, similarly cautious to PDM yet handling interactions better, with strong lane alignment likely inherited from its human-demonstration training data. Our hybrid planner outperforms the baselines in nearly all settings, combining PDM's safety with PPO's assertiveness. The substantial gap between the two splits across all planners further confirms that Interactive1k isolates a harder, more discriminative slice of the distribution.

Disaggregating the final score into the six core metrics in Fig.~\ref{fig:subscores_radar} exposes the underlying trade-offs. PDM's goal-completion gain on Random1k is most pronounced against Cautious traffic, where yielding agents block the ego and expose the limits of a non-negotiating planner. PPO instead exhibits a striking self-play artefact: its at-fault collision rate is lowest when paired with the PPO traffic agent and rises substantially against all others, peaking against the replay-based Expert; the policy thus overfits to behaviors seen during self-play and generalizes poorly to both reactive and non-reactive agents. PPO also scores low on all three behavioral metrics, indicating that the simple reward is insufficient for comfortable, lane-aware driving. SMART avoids this behavioral collapse but exhibits the highest off-road rate, a likely consequence of distribution shift in closed-loop rollouts, and its collision rate is markedly elevated against conservative traffic (IDM, Cautious, Mix), suggesting its learned interaction model is poorly calibrated to yielding behavior. Our hybrid planner successfully unifies PDM's safety with PPO's high goal completion, though behavioral metrics such as comfort and lane alignment still leave clear room for improvement. Qualitative results are shown in Appendix~\ref{sec:additional_res}.

The same pattern transfers to established benchmarks. Tab.~\ref{tab:val14_interplan} reports results on nuPlan Val14 and interPlan, where our PufferDrive--nuPlan interface enables the evaluation of an at-scale RL-trained planner in these settings. PDM dominates on the routine-driving Val14, while PPO surpasses PDM on the interaction-heavy interPlan and outperforms the imitation-based DiffusionPlanner~\citep{zheng2025diffusionplanner} by a large margin, consistent with the interpretation that imitation policies struggle on the out-of-distribution states arising in closed-loop rollouts.

\paragraph{Realism of traffic agents.}
Tab.~\ref{tab:realism_combined} reports the WOSAC 2024 realism score~\citep{montali2024wosac} with its kinematic, interactive, and map-based components, alongside minADE and ADE (per-subscore breakdown in Appendix Tab.~\ref{tab:realism_subscores}). SMART achieves the highest realism on both splits and approaches the ground-truth ceiling across all four metric groups, with its advantage most pronounced on the kinematic axis and trajectory errors, reflecting its behavior-cloning paradigm that inherits the empirical distribution of human driving. IDM remains competitive on the interactive and map-based axes but is limited kinematically as it only models longitudinal car-following, while PPO underperforms IDM on every axis, reflecting its simple reward.

Our reward-conditioned policies instead optimize a substantially richer reward, with the corresponding weights sampled and exposed to the network as conditioning variables. Although none matches SMART, since they optimize a parameterized reward rather than imitate humans, their realism meta scores are on par with IDM and consistently above PPO, indicating that the richer reward yields more humanlike behavior. The subscore decomposition reveals the intended behavioral separation. 

{\color{red}$\pi_\theta^{\text{aggr}}$} attains the highest linear acceleration and lowest trajectory error, consistent with its higher target speed approaching the average WOMD speed. {\color{purple}$\pi_\theta^{\text{norm}}$} shows a conservative balance between progress and safety, while {\color{green}$\pi_\theta^{\text{caut}}$} attains the highest collision and time-to-collision subscores, reflecting a more conservative driving style. $\pi_\theta^{\text{mix}}$ lies between the three modes throughout, showing that our reward-conditioned agents cover a well-separated, behaviorally meaningful slice of the driving distribution that complements imitation-based agents such as SMART.

\paragraph{Reward conditioning.}

\begin{wrapfigure}{r}{0.45\linewidth}
    \vspace{-1.2\baselineskip}
    \centering
    \includegraphics[width=\linewidth]{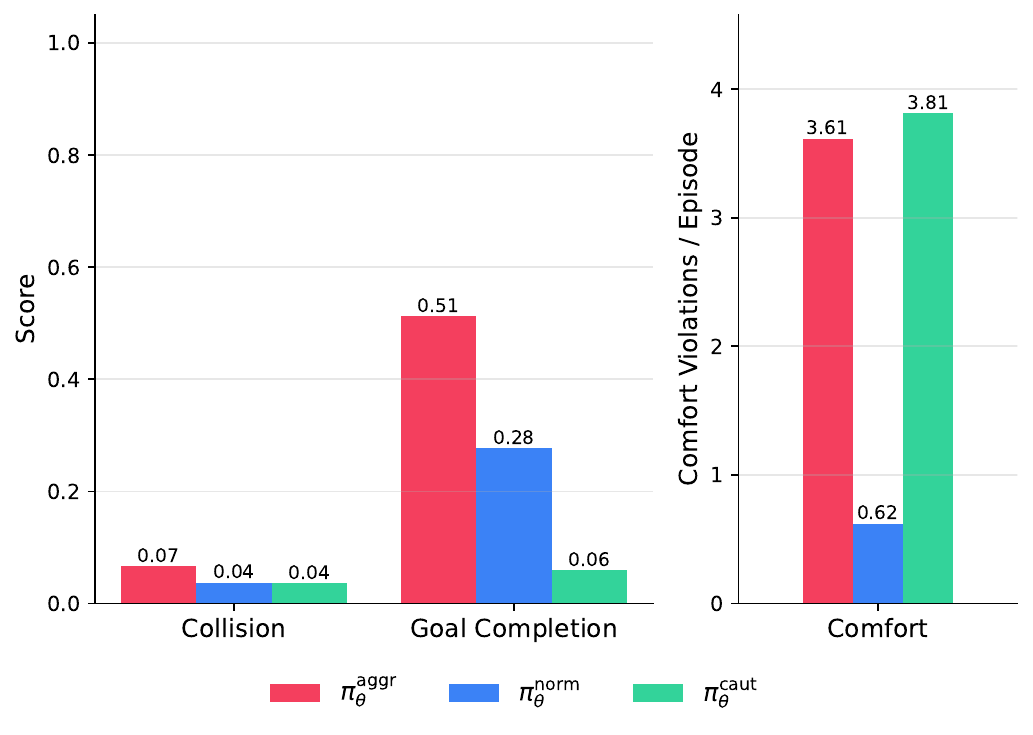}
    \caption{Collision, comfort, and goal-completion scores on Interactive1k for the three reward-conditioned policies ${\color{red}\pi_\theta^{\text{aggr}}}$, ${\color{purple}\pi_\theta^{\text{norm}}}$, and ${\color{green}\pi_\theta^{\text{caut}}}$, with the ego controlled by the conditioned policy and surrounding agents by IDM.}
    \label{fig:reward-conditioning}
    \vspace{-0.3\baselineskip}
\end{wrapfigure}

To verify that our traffic agents behave in accordance with their reward conditioning, we compare the three behavior modes across a set of complementary metrics. In each run, the ego vehicle is controlled by one of the three conditioned policies while the surrounding agents follow IDM. Fig.~\ref{fig:reward-conditioning} reports the collision, comfort, and goal-completion scores on the Interactive1k benchmark.
The results closely match the intended behavioral profiles. The aggressive policy ${\color{red}\pi_\theta^{\text{aggr}}}$ exhibits the highest collision rate, consistent with its preference for assertive maneuvers, while ${\color{purple}\pi_\theta^{\text{norm}}}$ and ${\color{green}\pi_\theta^{\text{caut}}}$ collide equally rarely. In terms of comfort, ${\color{purple}\pi_\theta^{\text{norm}}}$ incurs by far the fewest violations, whereas both ${\color{red}\pi_\theta^{\text{aggr}}}$ and ${\color{green}\pi_\theta^{\text{caut}}}$ show markedly more. For the aggressive policy, this stems from frequent lane changes and overtaking maneuvers, which induce high lateral accelerations. For the cautious policy, the low speed limit forces frequent decelerations, and its large time-to-collision threshold requires constant speed adjustments to keep a wide gap to the lead vehicle, further reducing comfort.

These results confirm that a single trained network can be steered into qualitatively different driving styles purely by reward conditioning, with each style trading off different driving characteristics in the expected manner. We sometimes observe qualitative unexpected behaviors, discussed further in Appendix~\ref{sec:additional_res}.

\paragraph{Scaling RL alone is not enough.}
\begin{wrapfigure}{r}{0.4\linewidth}
    \vspace{-1\baselineskip}
    \centering
    \includegraphics[width=\linewidth]{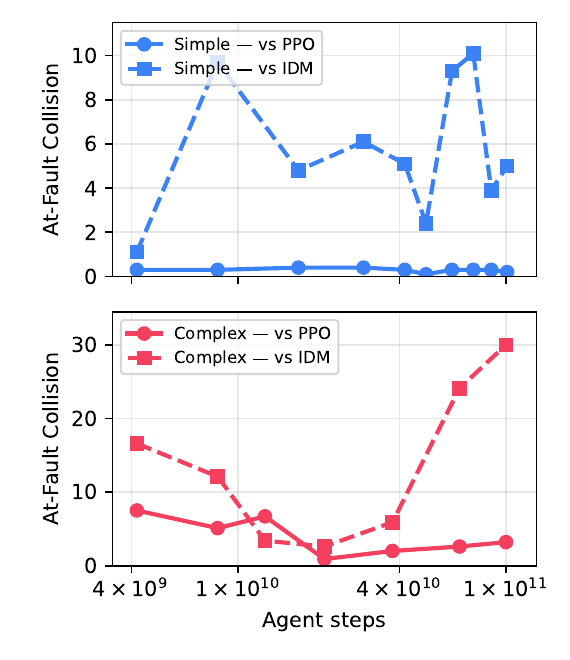}
    \caption{At-fault collision rate ($\%$) on the Interactive1k split for PPO planners trained via self-play under a simple (top) and complex (bottom) reward formulation, evaluated at intermediate checkpoints up to $10^{11}$ agent steps against PPO (solid) and IDM (dashed) traffic. Both reward formulations yield low collision rates against their own behavior but fail to generalize to IDM behavior.}
    \label{fig:scaling}
    \vspace{-1\baselineskip}
\end{wrapfigure}
Fig.~\ref{fig:scaling} reports the at-fault collision rate on the Interactive1k split for PPO planners trained via self-play under two reward formulations of differing complexity for $10^{11}$ agent steps. The simple reward covers only goal completion, collision, and off-road terms, whereas the complex reward additionally includes behavior shaping terms such as comfort and lane alignment, full definitions are provided in Sec.~\ref{sec:reward-baselines}. We evaluate intermediate checkpoints against both self-play (PPO) and IDM traffic. PPO trained with the simple reward yields a near-zero collision rate against PPO traffic across all checkpoints, indicating that the policy handles the behavior it was trained against reliably. When the same checkpoints are evaluated against IDM traffic, however, the collision rate remains an order of magnitude higher and shows no monotonic improvement throughout training. The complex reward exhibits the same pattern with a more pronounced gap. This asymmetry indicates that self-play training at scale in itself does not induce behavior generalization, and that scaling along the agent-step axis is insufficient to close the gap across traffic models.
GigaFlow offers a compelling solution to this generalization gap through reward conditioning, exposing the policy to a distribution over reward coefficients during training so that diverse behaviors are induced within self-play. The approach is highly effective and has been shown to yield robust, generalist policies. Two practical drawbacks remain, however. First, the method relies on massive training budgets: GigaFlow trains for ten days and accumulates $1.6$ billion  km of simulated driving. Second, once the network is trained, the design effort shifts from reward shaping to the search over the reward coefficient distribution itself. 

\section{Conclusion}
We introduced \underline{\textbf{BehaviorBench}}, addressing central limitations of prior planner evaluation through three contributions. First, an \textcolor{green}{\textbf{evaluation interface}} for at-scale RL planners, including a nuPlan interface for benchmarking RL-trained policies against established planners. Second, \textcolor{purple}{\textbf{scenario splits}} covering routine and highly interactive driving. Third, a \textcolor{red}{\textbf{diverse suite of traffic agents}} that exposes planner behavior far beyond the single-IDM setup of most benchmarks.

Our results reveal a finding that most prior benchmarks obscure: the strong performance reported is in large part an artefact of homogeneous evaluation conditions. Under diverse traffic behaviors and interactive scenarios, no single baseline simultaneously achieves safety, interaction handling, and behavioral smoothness. We further find that at-scale self-play RL planners overfit to the traffic behaviors they are trained against and degrade under shifts in traffic composition. Two complementary directions emerge as promising responses: concurrent work such as GigaFlow narrows the generalization gap through reward conditioning at extreme scale, while our hybrid planner shows that combining rule-based and learned components offer a lighter-weight alternative. BehaviorBench provides the first standardized testbed on which such directions can be systematically compared.
BehaviorBench thus moves the field beyond saturated single-traffic-model benchmarks toward stress-testing across diverse, interactive traffic, establishing a rigorous standard for the planning community and laying the foundation for future at-scale and RL-driven planning research.
Traffic lights and stop signs are not yet integrated, currently, we mitigate this by removing agents from the scene upon goal completion, with explicit traffic light handling left for future work. 


\clearpage
\begin{ack}
 This research was funded by the Deutsche Forschungsgemeinschaft (DFG, German Research Foundation) under grant number 539134284, through EFRE (FEIH\_2698644) and the state of Baden-Württemberg. \begin{center} \includegraphics[width=0.3\textwidth]{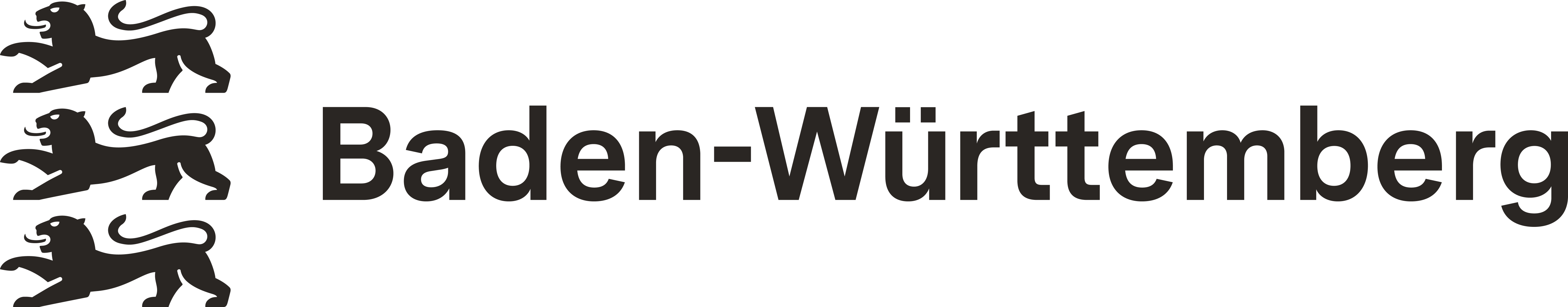} ~~~ \includegraphics[width=0.3\textwidth]{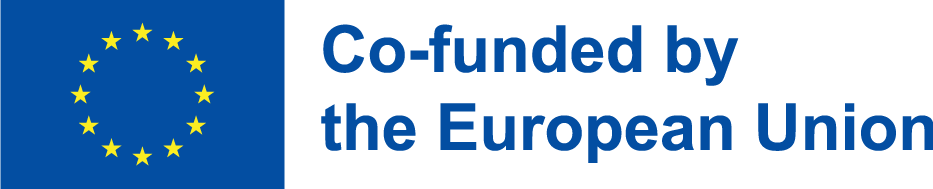} \end{center}
\end{ack}

\bibliographystyle{plainnat}
\bibliography{bib}  

@inproceedings{wu2024smart,
 author = {Wu, Wei and Feng, Xiaoxin and Gao, Ziyan and Kan, Yuheng},
 booktitle = {Advances in Neural Information Processing Systems},
 doi = {10.52202/079017-3622},
 editor = {A. Globerson and L. Mackey and D. Belgrave and A. Fan and U. Paquet and J. Tomczak and C. Zhang},
 pages = {114048--114071},
 publisher = {Curran Associates, Inc.},
 title = {SMART: Scalable Multi-agent Real-time Motion Generation via Next-token Prediction},
 volume = {37},
 year = {2024}
}

@InProceedings{Dauner2023CORL,
  title={Parting with Misconceptions about Learning-based Vehicle Motion Planning},
  author={Dauner, Daniel and Hallgarten, Marcel and Geiger, Andreas and Chitta, Kashyap},
  booktitle={Conference on Robot Learning (CoRL)},
  year={2023}
}

@inproceedings{kazemkhani2025gpudrive,
      title={GPUDrive: Data-driven, multi-agent driving simulation at 1 million FPS},
      author={Saman Kazemkhani and Aarav Pandya and Daphne Cornelisse and Brennan Shacklett and Eugene Vinitsky},
      booktitle={Proceedings of the International Conference on Learning Representations (ICLR)},
      year={2025},
      eprint={2408.01584},
      archivePrefix={arXiv},
      primaryClass={cs.AI},
}

@inproceedings{cusumanotowner2025gigaflow,
      title={Robust Autonomy Emerges from Self-Play}, 
      author={Marco Cusumano-Towner and David Hafner and Alex Hertzberg and Brody Huval and Aleksei Petrenko and Eugene Vinitsky and Erik Wijmans and Taylor Killian and Stuart Bowers and Ozan Sener and Philipp Krähenbühl and Vladlen Koltun},
      booktitle={International Conference on Machine Learning (ICML)},
      year={2025},
      eprint={2502.03349},
      archivePrefix={arXiv},
      primaryClass={cs.LG},
}

@misc{cornelisse2025pufferdrive,
      title={Building reliable sim driving agents by scaling self-play}, 
      author={Daphne Cornelisse and Aarav Pandya and Kevin Joseph and Joseph Suárez and Eugene Vinitsky},
      year={2025},
      eprint={2502.14706},
      archivePrefix={arXiv},
      primaryClass={cs.AI},
}

@inproceedings{nuplan,
      title={NuPlan: A closed-loop ML-based planning benchmark for autonomous vehicles}, 
      author={Holger Caesar and Juraj Kabzan and Kok Seang Tan and Whye Kit Fong and Eric Wolff and Alex Lang and Luke Fletcher and Oscar Beijbom and Sammy Omari},
      booktitle={CVPR ADP3 workshop},
      year={2022},
      eprint={2106.11810},
      archivePrefix={arXiv},
      primaryClass={cs.CV},
}

@inproceedings{karnchanachari2024nuplan,
  title={Towards Learning-Based Planning: The nuPlan Benchmark for Real-World Autonomous Driving},
  author={Karnchanachari, Napat and Geromichalos, Dimitris and Tan, Kok Seang and Li, Nanxiang and Eriksen, Christopher and Yaghoubi, Shakiba and Mehdipour, Noushin and Bernasconi, Gianmarco and Fong, Whye Kit and Guo, Yiluan and Caesar, Holger},
  booktitle={IEEE International Conference on Robotics and Automation (ICRA)},
  year={2024},
  doi={10.1109/ICRA57147.2024.10611484}
}

@inproceedings{Dosovitskiy17CARLA,
  title = {{CARLA}: {An} Open Urban Driving Simulator},
  author = {Alexey Dosovitskiy and German Ros and Felipe Codevilla and Antonio Lopez and Vladlen Koltun},
  booktitle = {Proceedings of the 1st Annual Conference on Robot Learning},
  pages = {1--16},
  year = {2017}
}

@inproceedings{waymax,
title={Waymax: An Accelerated, Data-Driven Simulator for Large-Scale Autonomous Driving Research},
author={Cole Gulino and Justin Fu and Wenjie Luo and George Tucker and Eli Bronstein and Yiren Lu and Jean Harb and Xinlei Pan and Yan Wang and Xiangyu Chen and John D. Co-Reyes and Rishabh Agarwal and Rebecca Roelofs and Yao Lu and Nico Montali and Paul Mougin and Zoey Yang and Brandyn White and Aleksandra Faust and Rowan McAllister and Dragomir Anguelov and Benjamin Sapp},
booktitle={Proceedings of the Neural Information Processing Systems Track on Datasets and Benchmarks},
year={2023}
}

@article{Schulman2017ppo,
  title={Proximal Policy Optimization Algorithms},
  author={John Schulman and Filip Wolski and Prafulla Dhariwal and Alec Radford and Oleg Klimov},
  journal={ArXiv},
  year={2017},
  volume={abs/1707.06347},
}

@inproceedings{waymo_dataset,
  title={Large scale interactive motion forecasting for autonomous driving: The waymo open motion dataset},
  author={Ettinger, Scott and Cheng, Shuyang and Caine, Benjamin and Liu, Chenxi and Zhao, Hang and Pradhan, Sabeek and Chai, Yuning and Sapp, Ben and Qi, Charles R and Zhou, Yin and others},
  booktitle={Proceedings of the IEEE/CVF International Conference on Computer Vision},
  pages={9710--9719},
  year={2021}
}

@article{Treiber2000idm,
   title={Congested traffic states in empirical observations and microscopic simulations},
   volume={62},
   ISSN={1095-3787},
   DOI={10.1103/physreve.62.1805},
   number={2},
   journal={Physical Review E},
   publisher={American Physical Society (APS)},
   author={Treiber, Martin and Hennecke, Ansgar and Helbing, Dirk},
   year={2000},
   month=aug, pages={1805–1824} }

@misc{distelzweig2025spdm,
      title={Perfect Prediction or Plenty of Proposals? What Matters Most in Planning for Autonomous Driving}, 
      author={Aron Distelzweig and Faris Janjoš and Oliver Scheel and Sirish Reddy Varra and Raghu Rajan and Joschka Boedecker},
      year={2025},
      eprint={2510.15505},
      archivePrefix={arXiv},
      primaryClass={cs.RO}, 
}

@inproceedings{hallgarten2024interplan,
author = {Hallgarten, Marcel and Zapata, Julian and Stoll, Martin and Renz, Katrin and Zell, Andreas},
year = {2024},
month = {10},
pages = {5388-5395},
title = {Can Vehicle Motion Planning Generalize to Realistic Long-tail Scenarios?},
booktitle={IEEE International Conference on Intelligent Robots and Systems (IROS)},
doi = {10.1109/IROS58592.2024.10803052}
}

@misc{montali2024wosac,
      title={The Waymo Open Sim Agents Challenge}, 
      author={Nico Montali and John Lambert and Paul Mougin and Alex Kuefler and Nick Rhinehart and Michelle Li and Cole Gulino and Tristan Emrich and Zoey Yang and Shimon Whiteson and Brandyn White and Dragomir Anguelov},
      year={2023},
      eprint={2305.12032},
      archivePrefix={arXiv},
      primaryClass={cs.CV},
      url={https://arxiv.org/abs/2305.12032}, 
}

@inproceedings{huang2024dtpp,
  title={DTPP: Differentiable Joint Conditional Prediction and Cost Evaluation for Tree Policy Planning in Autonomous Driving},
  author={Huang, Zhiyu and Karkus, Peter and Ivanovic, Boris and Chen, Yuxiao and Pavone, Marco and Lv, Chen},
  booktitle={2024 IEEE International Conference on Robotics and Automation (ICRA)},
  pages={6806--6812},
  year={2024}
}

@inproceedings{cheng2024plantf,
author = {Cheng, Jie and Chen, Yingbing and Mei, Xiaodong and Yang, Bowen and Li, Bo and Liu, Ming},
year = {2024},
month = {05},
pages = {14123-14130},
title = {Rethinking Imitation-based Planners for Autonomous Driving},
booktitle={IEEE International Conference on Robotics and Automation (ICRA)},
doi = {10.1109/ICRA57147.2024.10611364}
}

@InProceedings{Jaeger2025CoRL, 
	author = {Bernhard Jaeger and Daniel Dauner and Jens Beißwenger and Simon Gerstenecker and Kashyap Chitta and Andreas Geiger}, 
	title = {CaRL: Learning Scalable Planning Policies with Simple Rewards}, 
	booktitle = {Proc. of the Conf. on Robot Learning (CoRL)}, 
	year = {2025}, 
}

@inproceedings{
zheng2025diffusionplanner,
title={Diffusion-Based Planning for Autonomous Driving with Flexible Guidance},
author={Yinan Zheng and Ruiming Liang and Kexin ZHENG and Jinliang Zheng and Liyuan Mao and Jianxiong Li and Weihao Gu and Rui Ai and Shengbo Eben Li and Xianyuan Zhan and Jingjing Liu},
booktitle={The Thirteenth International Conference on Learning Representations},
year={2025},
}

@inproceedings{chang2019argoverse,
  title     = {{Argoverse}: {3D} Tracking and Forecasting with Rich Maps},
  author    = {Chang, Ming-Fang and Lambert, John and Sangkloy, Patsorn and Singh, Jagjeet and Bak, Slawomir and Hartnett, Andrew and Wang, De and Carr, Peter and Lucey, Simon and Ramanan, Deva and Hays, James},
  booktitle = {Proceedings of the {IEEE/CVF} Conference on Computer Vision and Pattern Recognition ({CVPR})},
  year      = {2019},
  doi       = {10.1109/CVPR.2019.00895}
}

@article{charraut2025vmax,
  title={V-Max: A Reinforcement Learning Framework for Autonomous Driving},
  author={Charraut, Valentin and Doulazmi, Wa{\"e}l and Tournaire, Thomas and Buhet, Thibault},
  journal={arXiv preprint arXiv:2503.08388},
  year={2025}
}

@inproceedings{yin2024regents,
  title={Regents: Real-world safety-critical driving scenario generation made stable},
  author={Yin, Yuan and Khayatan, Pegah and Zablocki, {\'E}loi and Boulch, Alexandre and Cord, Matthieu},
  booktitle={European Conference on Computer Vision},
  pages={262--276},
  year={2024},
  organization={Springer}
}

@article{li2023scenarionet,
  title={Scenarionet: Open-source platform for large-scale traffic scenario simulation and modeling},
  author={Li, Quanyi and Peng, Zhenghao Mark and Feng, Lan and Liu, Zhizheng and Duan, Chenda and Mo, Wenjie and Zhou, Bolei},
  journal={Advances in neural information processing systems},
  volume={36},
  pages={3894--3920},
  year={2023}
}

@article{jaeger2025carl,
  title={Carl: Learning scalable planning policies with simple rewards},
  author={Jaeger, Bernhard and Dauner, Daniel and Bei{\ss}wenger, Jens and Gerstenecker, Simon and Chitta, Kashyap and Geiger, Andreas},
  journal={arXiv preprint arXiv:2504.17838},
  year={2025}
}

@misc{suarez2024pufferlib,
      title={PufferLib: Making Reinforcement Learning Libraries and Environments Play Nice}, 
      author={Joseph Suarez},
      year={2024},
      eprint={2406.12905},
      archivePrefix={arXiv},
      primaryClass={cs.LG},
}

@inproceedings{caesar2020nuscenes,
  title     = {nuScenes: A Multimodal Dataset for Autonomous Driving},
  author    = {Caesar, Holger and Bankiti, Varun and Lang, Alex H. and Vora, Sourabh and Liong, Venice Erin and Xu, Qiang and Krishnan, Anush and Pan, Yu and Baldan, Giancarlo and Beijbom, Oscar},
  booktitle = {Proceedings of the IEEE/CVF Conference on Computer Vision and Pattern Recognition (CVPR)},
  year      = {2020}
}

@article{li2022metadrive,
  title={Metadrive: Composing diverse driving scenarios for generalizable reinforcement learning},
  author={Li, Quanyi and Peng, Zhenghao and Feng, Lan and Zhang, Qihang and Xue, Zhenghai and Zhou, Bolei},
  journal={IEEE Transactions on Pattern Analysis and Machine Intelligence},
  year={2022}
}


\clearpage
\appendix
\section{Related Work}
\label{sec:related_work}
In this section, we review datasets and benchmarks that support closed-loop simulation, distinguishing our scope from perception-centric prediction benchmarks like nuScenes~\citep{caesar2020nuscenes} or Argoverse~\citep{chang2019argoverse}. We include the Waymo Open Motion Dataset (WOMD)~\citep{waymo_dataset} due to its essential role in validating simulation fidelity via the WOSAC framework, which plays an important role in this work. We continue discussing simulation frameworks and their capabilities in the last part of this section. A detailed feature-based comparison of these frameworks is presented in Table \ref{tab:simulator_comparison}.

\subsection{Datasets and Benchmarks}
nuPlan~\citep{karnchanachari2024nuplan} serves as the primary large-scale dataset and benchmark for closed-loop planning, comprising 1{,}282 hours of driving data from four cities with high-fidelity object tracks, traffic lights, and over 70 scenario types. Several benchmarks have been proposed on top of nuPlan to evaluate closed-loop planners, each addressing different shortcomings but introducing new limitations.

The widely adopted \textbf{Val14} benchmark~\citep{Dauner2023CORL} provides a standardized subset of 1{,}118 nuPlan scenarios evaluated using the Closed-Loop Score (CLS), but predominantly covers basic driving situations.
To address this, \textbf{Test14-Hard}~\citep{cheng2024plantf} selects the 20 hardest nuPlan scenarios per type based on the performance of PDM-Closed~\citep{Dauner2023CORL}, a strong rule-based planner, however, this selection criterion introduces a planner-specific bias rather than principled long-tail sampling.

Going beyond scenario selection, \textbf{interPlan}~\citep{hallgarten2024interplan} augments nuPlan with 80 hand-designed interactive scenarios across eight categories and introduces conservative and assertive IDM traffic agent policies by using different IDM parameters. While it reveals critical planner shortcomings, its agent variants remain modifications of IDM's reaction timing rather than fundamentally different behavioral models, and hand-crafted augmentation at this scale does not cover the diversity required for comprehensive evaluation. We address the agent diversity limitation in Subsection \ref{subsec:agents}.

WOSAC~\citep{montali2024wosac} evaluates simulation fidelity by measuring the distributional realism of 8-second rollouts against the WOMD, which provides over 570 hours of interactive human driving data across six U.S.\ cities. We adopt the WOSAC~2024 metrics in our framework to assess the realism of our traffic agents (Tab.~\ref{tab:realism_combined}).

\subsection{Simulation Frameworks}
\label{sec:sim_frameworks}

In the following section, we contextualize BehaviorBench by comparing it against prominent simulation frameworks--specifically CARLA\,\citep{Dosovitskiy17CARLA}, MetaDrive~\citep{li2022metadrive}, Waymax~\citep{waymax}, GPUDrive~\citep{kazemkhani2025gpudrive}, PufferDrive~\citep{cornelisse2025pufferdrive}, nuPlan~\citep{nuplan}, and V-Max~\citep{charraut2025vmax}. Our analysis evaluates these platforms based on their supported datasets and benchmarks, traffic agent modeling and multi-agent capabilities, simulation throughput, and the integration of advanced simulator features. A detailed comparison of simulators is provided in Appendix Tab.~\ref{tab:simulator_comparison}. This comparison provides the necessary context to understand how BehaviorBench relates to existing tools in terms of functionality and performance across these core categories. 

\paragraph{Supported Datasets and Benchmarks}
The field of autonomous driving simulation ranges from localized, synthetic environments to massive-scale, real-world data integration to enhance model generalizability. While legacy simulators like CARLA are often restricted to low-diversity synthetic datasets, modern frameworks such as Waymax, GPUDrive, and PufferDrive prioritize the Waymo Open Motion Dataset (WOMD), frequently extending it with the Waymo Open Sim Agents Challenge (WOSAC) for more robust behavioral modeling. Similarly, nuPlan employs a multi-tiered evaluation structure, utilizing specialized subsets like Val14, Test14-Hard, and interPlan to stress-test agents under diverse conditions.  
To overcome inherent data limitations, MetaDrive builds on the ScenarioNet~\citep{li2023scenarionet} framework, which provides access to a diverse set of driving datasets, including Waymo, nuPlan, and Argoverse. In addition to these real-world sources, MetaDrive supports synthetically generated data, ranging from safety-critical traffic scenarios to procedurally generated road networks. Both V-Max and BehaviorBench utilize the ScenarioMax framework, an extension of ScenarioNet~\citep{li2023scenarionet}. This enables seamless scaling across heterogeneous datasets and facilitates a dataset-agnostic training environment that surpasses the native constraints of earlier implementations. The primary distinction between these high-scale simulators lies in their evaluative focus: while V-Max assesses agent robustness against adversarial actors via ReGentS~\citep{yin2024regents}, BehaviorBench expands the benchmarking of complex behaviors by introducing two additional evaluation datasets.

\paragraph{Traffic agent modeling and Multi-Agent Support}
Effective agent steering is essential for autonomous driving research, requiring a mix of non-reactive and reactive behaviors to ensure robust training and evaluation. MetaDrive supports multiple execution modes for real-world data, including log-replay, rule-based simulation and learning-based control. In contrast, simulators such as nuPlan and CARLA typically focus on single-agent training where the ego-vehicle interacts with rule-based traffic.  While V-Max utilizes log-replay for training, it offers a diverse preimplemented evaluation suite including imitation learning, RL, and rule-based policies like IDM and PDM. Waymax, GPUDrive, and PufferDrive all enable multi-agent training, however, they often rely on a self-play paradigm where a single model controls multiple objects. This reliance on self-play often results in overfitting to the model’s specific behavioral patterns. GigaFlow~\citep{cusumanotowner2025gigaflow}—the closed-source predecessor to GPUDrive and PufferDrive—addresses this limitation by employing reward conditioning to foster greater behavioral diversity. Waymax supports user-defined agent behaviors with pre-implemented log-replay and IDM agents. BehaviorBench supports PPO-only self-play and leverages reward conditioning to generate a theoretically infinite spectrum of behavioral styles (e.g., cautious to aggressive). Furthermore, it allows for the inclusion of log-replay and IDM-based agents via adjustable fractions. 

\newcommand{\cmark}{\checkmark}
\newcommand{\xmark}{$\times$}

\begin{table}[t]
\centering
\resizebox{\textwidth}{!}{%
\begin{tabular}{ll cc || ccc | c || ccccc}
\textbf{Simulator} &
\textbf{Dataset / Map} &
\textbf{Diff.} &
\textbf{Scale} &
\rotatebox{90}{\textbf{Log replay}} &
\rotatebox{90}{\textbf{Rule-based}} &
\rotatebox{90}{\textbf{Learning-based}} &
\rotatebox{90}{\textbf{Self-play}} &
\rotatebox{90}{\textbf{CARLA LB}} &
\rotatebox{90}{\textbf{Val14}} &
\rotatebox{90}{\textbf{Test14-Hard}} &
\rotatebox{90}{\textbf{interPlan}} &
\rotatebox{90}{\textbf{WOSAC}} \\
\midrule
CARLA~\citep{Dosovitskiy17CARLA}
  & Synthetic
  & \xmark & \texttt{-}
  & \xmark & \cmark & \xmark & \xmark
  & \cmark & \xmark & \xmark & \xmark & \xmark \\

MetaDrive~\citep{li2022metadrive} & Synthetic + WOMD + nuPlan \dag
  & \xmark & \texttt{-}
  & \cmark & \cmark & \cmark & \cmark
  & \xmark & \xmark & \xmark  & \xmark & \xmark \\
\midrule

nuPlan~\citep{karnchanachari2024nuplan}
  & nuPlan
  & \xmark & \texttt{-}
  & \cmark & \cmark & \xmark & \xmark
  & \xmark & \cmark & \cmark & \cmark & \xmark \\
\midrule
Waymax~\citep{waymax}
  & WOMD
  & \cmark & $\circ$
  & \cmark & \cmark & \cmark & \xmark
  & \xmark & \xmark & \xmark & \xmark & \cmark \\
V-Max~\citep{charraut2025vmax}
  & WOMD + nuPlan
  & \cmark & $\circ$
  & \cmark & \cmark & \cmark & \xmark
  & \xmark & \cmark & \cmark & \xmark & \cmark \\
\midrule


GPUDrive~\citep{kazemkhani2025gpudrive}
  & WOMD
  & \xmark & $\circ$
  & \cmark & \xmark & \cmark & \cmark
  & \xmark & \xmark & \xmark & \xmark & \xmark \\
PufferDrive~\citep{cornelisse2025pufferdrive}
  & WOMD
  & \xmark & \texttt{+}
  & \xmark & \xmark & \cmark & \cmark
  & \xmark & \xmark & \xmark & \xmark & \cmark \\
\textbf{BehaviorBench (Ours)}
  & \textbf{WOMD + nuPlan}
  & \xmark & \texttt{+}
  & \cmark & \cmark & \cmark & \cmark
  & \xmark & \cmark & \cmark & \cmark & \cmark \\
\bottomrule
\end{tabular}%
}
\caption{%
Comparison of open-source autonomous driving simulators. 
\emph{Diff.}: gradients propagate through simulation dynamics.
\emph{Scale}: Relative simulation throughput. Categorized as
slower (\,\texttt{-}\,),
comparable (\,\scalebox{0.8}{$\circ$}\,) or
faster (\,\texttt{+}\,) relative to the GPUDrive baseline ~\citep{kazemkhani2025gpudrive}).
Agent-model columns indicate which traffic-agent models are supported during evaluation (Log replay, Rule-based, Learning-based) and whether Self-play can be used during training. 
Benchmark columns indicate which standardised evaluation suites are supported.
\dag~indicates support for datasets included in ScenarioNet~\citep{li2023scenarionet}.
}
\label{tab:simulator_comparison}
\end{table}

\paragraph{Simulation Scale}
In the landscape of high-throughput multi-agent simulation, a distinct hierarchy of computational performance has emerged among leading frameworks. GPUDrive demonstrates a substantial efficiency advantage, outperforming Waymax~\citep{kazemkhani2025gpudrive} in throughput. Empirical evidence from the V-Max literature demonstrates that runtimes are largely commensurate with the Waymax baseline, potentially even slightly lower, while simultaneously claiming superior computational performance over nuPlan and MetaDrive~\citep{charraut2025vmax}. This performance gap is further contextualized by CaRL~\citep{jaeger2025carl}, which, despite providing significant acceleration for CARLA, notes that its performance still does not fully reach the level of nuPlan. PufferDrive exceeds GPUDrive’s benchmarks in execution speed, offering a highly optimized solution for large-scale tasks. By building on this architecture, Behavior Bench inherits these gains to provide high-throughput simulation. 

\paragraph{Advanced Simulator Features}
Realistic partial observability and the inclusion of pedestrians and cyclists are critical for simulating real-world driving complexity. While CARLA provides partial observability via camera inputs, Waymax, and V-Max employ LiDAR-based masking where objects physically obstruct line-of-sight. In contrast, GPUDrive, PufferDrive, and nuPlan assume full observability. Regarding agent diversity, most frameworks support pedestrians and cyclists, though they are notably absent in PufferDrive and MetaDrive. Additionally, Waymax stands as the only platform offering native differentiability; while V-Max theoretically shares this capability through its Waymax integration, it is not utilized to the best of our knowledge. While BehaviorBench supports the integration of pedestrians and cyclists, it currently operates under the assumption of full observability.


\section{Evaluation Metrics}
\label{sec:eval-metrics}

Each scenario is summarized by six scalars: three continuous \emph{behavior scores} $\in [0, 1]$ (comfort, lane alignment, lane center) and three binary \emph{outcome indicators} $\in \{0, 1\}$ (goal, off-road, at-fault collision). Let $T$ be the number of steps the ego was alive.

\subsection{Behavior scores}

\paragraph{Comfort ($S_{\text{cmf}}$).} At each active step the ego can
incur up to three comfort violations
\begin{equation}
v^{(t)} = \mathbb{I}_{|a^{(t)}_{\text{long}}| > 3} + \mathbb{I}_{|a^{(t)}_{\text{lat}}| > 3} + \mathbb{I}_{\max(|\dot a^{(t)}_{\text{long}}|, |\dot a^{(t)}_{\text{lat}}|) > 5} \in \{0,1,2,3\},
\end{equation}
with accelerations in $\text{m}/\text{s}^2$ and jerks in
$\text{m}/\text{s}^3$. Summing $V = \sum_{t=1}^{T} v^{(t)} \in [0, 3T]$
gives
\begin{equation}
S_{\text{cmf}} \;=\; 1 - \min\!\left(\frac{V}{3\,T},\; 1\right).
\end{equation}
A fully comfortable scenario scores $1$, saturating all three violation types every step scores $0$.

\paragraph{Lane alignment ($S_{\text{align}}$).} Fraction of active steps at which the ego's heading is within a fixed angular tolerance $\tau$ of its nearest lane segment. With $\theta_{\text{ego}}^{(t)}$ the ego heading and $\theta_{\text{lane}}^{(t)}$ the heading of the nearest lane segment:
\begin{equation}
S_{\text{align}} \;=\; \frac{1}{T}\sum_{t=1}^{T} \mathbb{I}\!\left[\bigl|\theta_{\text{ego}}^{(t)} - \theta_{\text{lane}}^{(t)}\bigr| \,<\, \tau\right],
\qquad \tau = \tfrac{\pi}{12}.
\end{equation}
\paragraph{Lane center ($S_{\text{ctr}}$).} Mean absolute lateral distance $d^{(t)}$ from the ego to the closest lane centerline, averaged over active steps and normalized by $L_{\text{ref}} = 2.0\,\text{m}$:
\begin{equation}
\bar d = \frac{1}{T}\sum_{t=1}^{T} d^{(t)}, \qquad
S_{\text{ctr}} = 1 - \min\!\left(\frac{\bar d}{L_{\text{ref}}},\; 1\right).
\end{equation}

\subsection{Outcome indicators}

\paragraph{Goal completion ($S_{\text{goal}}$).} $S_{\text{goal}} = 1$ iff the ego reaches its scenario goal (within $r_{\text{goal}} = 2\,\text{m}$ and below a certain speed threshold), $0$ otherwise.

\paragraph{Off-road ($S_{\text{off}}$).} $S_{\text{off}} = 1$ iff the ego crosses the road edge at any step, $0$ otherwise.

\paragraph{At-fault collision ($S_{\text{col}}$).} $S_{\text{col}} = 1$ iff the ego caused a collision (Sec.~\ref{sec:at-fault}), $0$ otherwise.

\subsection{At-fault classification}
\label{sec:at-fault}

Following nuPlan, we distinguish several categories of at fault collisions. The classification comprises three criteria: whether the agent is stationary (speed below $v_{\text{stop}} = 0.1\,\text{m/s}$), where the other agent sits relative to the ego heading, and whether the ego is approaching it. We refer to an agent as ahead when it lies within a $60^{\circ}$ frontal cone of the ego, and as behind when it lies in the rear $30^{\circ}$ sector, the remaining angles are treated as lateral. Approaching means the gap between the two agents is closing at more than $0.5\,\text{m/s}$.

If the ego is stationary at the moment of contact (\textsc{stopped-ego}), the collision is not at fault. If instead the other agent is stationary (\textsc{stopped-track}), the ego must have driven into it and is classified as at fault. Driving into an agent ahead (\textsc{active-front}) is classified as at fault, while an agent driving into the ego from behind (\textsc{active-rear}) is not. The remaining lateral (\textsc{active-lateral}) and rear (\textsc{active-rear}) cases are only at fault when the ego's trajectory is consistent with a lane change, identified if it's overlapping with two or more lane polylines and a lateral drift of more than $0.3\,\text{m}$ relative to its heading. Both conditions must hold, which avoids false attributions on curved or narrow roads. Contacts with vulnerable road users are always classified as at fault. Table~\ref{tab:collision-types} summaries the resulting categories.

\begin{table}[h]
\centering
\small
\begin{tabular}{@{}llll@{}}
\toprule
Type & Condition & Interpretation & At fault \\
\midrule
\textsc{stopped-ego}    & ego stationary                       & ego is standing still         & no \\
\textsc{stopped-track}  & other agent stationary               & ego drives into stopped agent & yes \\
\textsc{active-rear}    & other agent behind ego               & contact comes from behind     & only if lane change \\
\textsc{active-front}   & other agent ahead, ego approaching   & ego closes in on agent ahead  & yes \\
\textsc{active-lateral} & otherwise                            & side-on contact               & only if lane change \\
\bottomrule
\end{tabular}
\caption{Collision categories and resulting fault assignment.}
\label{tab:collision-types}
\end{table}
\section{Interactive1k}
\label{sec:i1k}
In this section, we describe the components of $S_{\text{int}}$ in more detail:
\begin{itemize}
\item \textbf{Trajectory crossings} ($c_{\text{cross}}$): number of unique agents whose ground-truth trajectories geometrically intersect with the ego's trajectory, indicating potential interaction.
\item \textbf{Proximity-weighted acceleration} ($c_{\text{accel}}$): sum of ego acceleration changes, each weighted by the inverse distance to the nearest agent, capturing reactive braking or acceleration maneuvers caused by other traffic participants.
\item \textbf{Proximity-weighted steering} ($c_{\text{steer}}$): analogous to the above for steering changes, capturing lateral evasive maneuvers near other agents.
\item \textbf{TTC-critical steps} ($c_{\text{ttc}}$): number of simulation steps where the ego's time-to-collision with any approaching agent falls below 3\,s, measuring sustained exposure to near-collision situations.
\item \textbf{Nearby agents} ($c_{\text{agents}}$): number of agents within a 40\,m radius of the ego at $t=0$, reflecting initial scene density.
\item \textbf{Goal distance} ($c_{\text{goal}}$): distance to the ego's goal, favoring longer scenarios that require sustained planning.
\end{itemize}
We provide the weights $w_i$ and the normalization thresholds $\tau_i$ in Tab.~\ref{tab:score_components}. The normalization thresholds $\tau_i$ were set to the 75th--90th percentile of each component's distribution, so that typical scenarios score well below 1.0 while genuinely interactive ones saturate the individual terms. Scenarios are excluded if the ego goal distance is below 10\,m or fewer than 3 agents are present.

\begin{table}[ht]
    \centering
    \begin{tabular}{llcc}
        \toprule
        Comp. & Description & $w_i$ & $\tau_i$ \\
        \midrule
        $c_{\text{cross}}$  & Traj.\ crossings      & 0.30 & 4 \\
        $c_{\text{accel}}$  & Ego accel.\ changes   & 0.15 & 60 \\
        $c_{\text{steer}}$  & Ego steering changes  & 0.15 & 0.1 \\
        $c_{\text{ttc}}$    & TTC $< 3$\,s          & 0.20 & 60 \\
        $c_{\text{agents}}$ & Agents $< 40$\,m      & 0.10 & 10 \\
        $c_{\text{goal}}$   & Goal distance         & 0.10 & 100 \\
        \bottomrule
    \end{tabular}
    \caption{Components of the interactivity score $S_{\text{int}}$.}
    \label{tab:score_components}
\end{table}
\section{Network Architecture}
\label{sec:architecture}
Both the unconditional and reward-conditioned policies share the same encoder architecture.  A per-agent observation is partitioned into three input groups: the ego state, up to $M_P$ partner objects, and up to $M_R$ road objects. The conditioned network additionally receives a fourth group containing the ten reward-conditioning coefficients $c$, which we describe in Sec.~\ref{sec:reward-conditioning}. We train both networks for $10^{11}$ agent steps on $8$ NVIDIA L40S GPUs (48\,GB each), taking approximately $20$\,h at a throughput of $1.7$\,M steps per second.

\paragraph{Unconditional policy ($\pi_\theta$).}
The three branch embeddings are concatenated and projected by a shared MLP into a single latent $z \in \mathbb{R}^{H_{\text{sh}}}$:
\begin{equation}
z \;=\; f_{\text{emb}}\bigl(\, [\,h_{\text{ego}};\, h_{\text{road}};\, h_{\text{part}}\,]\,\bigr).
\label{eq:embed-base}
\end{equation}
The latent is then passed through a single-layer LSTM that integrates information across time, and the recurrent state feeds a categorical actor and a scalar critic:
\begin{equation}
h_t = \mathrm{LSTM}(z_t, h_{t-1}), \qquad
\pi_\theta(a \mid o) = \mathrm{Cat}\!\bigl(\mathrm{Actor}(h_t)\bigr), \qquad
V_\theta(o) = \mathrm{Critic}(h_t).
\end{equation}
The discrete action head produces $|\mathcal{A}|=91$ logits corresponding to the Cartesian product of 7 longitudinal accelerations and 13 steering angles.

\paragraph{Reward-conditioned policy ($\pi_\theta^{c}$).}
The conditioned variant adds a fourth encoder branch that embeds the ten-dimensional conditioning vector $c$, and extends the concatenation into the shared projection accordingly:
\begin{equation}
z^{c} \;=\; f_{\text{emb}}\bigl(\, [\,h_{\text{ego}};\, h_{\text{road}};\, h_{\text{part}};\, h_{c}\,]\,\bigr).
\label{eq:embed-cond}
\end{equation}
The LSTM, actor and critic keep the structure of Eq.~\eqref{eq:embed-base}, but the conditioned agents are trained under the jerk dynamics model and therefore act in a different space: the head produces $|\mathcal{A}^{c}|=99$ logits over the Cartesian product of 11 longitudinal and 9 lateral jerk values. By re-sampling $c \sim p(c)$ at the start of every training episode, the same network learns a continuous family of driving styles, at evaluation, pinning $c$ to a specific point of the sampling distribution (Tab.~\ref{tab:reward-conditioning}) instantiates the corresponding behavior.

\begin{table}[h]
\centering
\small
\begin{tabular}{@{}lcc@{}}
\toprule
 & \textbf{Unconditional} & \textbf{Conditioned} \\
\midrule
Encoder branches        & ego, road, partner      & + c\_reward          \\
Shared embedding input  & $[h_{\text{ego}}; h_{\text{road}}; h_{\text{part}}]$
                        & $[h_{\text{ego}}; h_{\text{road}}; h_{\text{part}}; h_{c}]$ \\
Recurrent core          & LSTM                    & LSTM               \\
Actor / Critic          & discrete 91 / scalar    & discrete 99 / scalar \\
\#Parameters            & 614k    & 637k \\
\bottomrule
\end{tabular}
\caption{Side-by-side summary. The conditioned variant differs in the added $c_\text{reward}$ branch and in its output head, which operates in a jerk-based action space (11 longitudinal $\times$ 9 lateral jerk values) instead of the acceleration--steering space of the unconditional policy.}
\label{tab:arch-summary}
\end{table}

\section{Reward Function}
\label{sec:reward-baselines}
Following GigaFlow~\citep{cusumanotowner2025gigaflow}, we use a similar per-step reward
\begin{equation}
\begin{split}
R_{t} \;=\;{}& R_{\text{goal}} + R_{\text{collision}} + R_{\text{off-road}}
            + R_{\text{speed}} + R_{\text{TTC}} + R_{\text{comfort}} \\
            & + R_{\text{l-align}} + R_{\text{l-center}} + R_{\text{velocity}}
            + R_{\text{reverse}} + R_{\text{timestep}} .
\end{split}
\end{equation}
with each term definition provided in Tab.~\ref{tab:reward-conditioning}. We train two unconditioned networks via PPO that share an identical architecture and differ only in the per-step reward. The simple variant assigns (negative) rewards only for collisions, off-road, and goal completion, whereas the complex variant follows \cite{cusumanotowner2025gigaflow} and includes nearly all terms. The fixed coefficients for the simple and complex version are provided in Tab.~\ref{tab:simple-vs-complex}.

\begin{table}[ht]
  \centering
  \small
  \begin{tabular}{lcc}
    \toprule
    Coefficient & Simple & Complex \\
    \midrule
    $\delta_{\text{goal}}$       & $2.0$ & $2.0$    \\
    $\alpha_{\text{collision}}$  & $0.5$ & $3.0$    \\
    $\alpha_{\text{boundary}}$   & $0.5$ & $3.0$    \\
    $v_{\text{limit}}$           & $0$   & $0$      \\
    $t_{\text{TTC}}$             & $0$   & $0$      \\
    $\alpha_{\text{comfort}}$    & $0$   & $0.05$   \\
    $\alpha_{\text{l-align}}$    & $0$   & $0.025$  \\
    $\alpha_{\text{vel-align}}$  & $0$   & $1.0$    \\
    $\alpha_{\text{l-center}}$   & $0$   & $0.0038$ \\
    $\alpha_{\text{center-bias}}$& $0$   & $0$      \\
    $\alpha_{\text{velocity}}$   & $0$   & $0.0025$ \\
    $\alpha_{\text{reverse}}$    & $0$   & $0.005$  \\
    $\alpha_{\text{timestep}}$   & $0$   & $2.5\!\times\!10^{-5}$ \\
    \bottomrule
  \end{tabular}
  \caption{Fixed reward coefficients for the two non-conditioned baselines. $v_{\text{limit}} = 0$ / $t_{\text{TTC}} = 0$ disable the respective threshold penalties.}
  \label{tab:simple-vs-complex}
\end{table}

\section{Reward Conditioning}
\label{sec:reward-conditioning}
In the per-step reward $R_t$ each term is weighted by a coefficient $\alpha$. We train a conditioned network via PPO, that observes the coefficients as part of the input. The coefficients are resampled uniformly at the start of every training episode, conditioning the agent on a driving style. In evaluation, the coefficients can be held fixed to instantiate a specific driving behavior.

At every episode reset, each agent samples a ten-dimensional conditioning vector
\begin{equation}
c = (v_{\text{limit}}, \alpha_{\text{collision}},
\alpha_{\text{boundary}}, \alpha_{\text{comfort}},
\alpha_{\text{l-align}}, \alpha_{\text{vel-align}},
\alpha_{\text{l-center}}, \alpha_{\text{center-bias}},
\alpha_{\text{reverse}}, t_{\text{TTC}})
\end{equation}
from the training distribution in Tab.~\ref{tab:reward-conditioning}. Each component is normalized to $[-1, 1]$ using the sampling range and appended to the observation. Note that two components condition the behavior through \emph{thresholds} rather than weights: exceeding the speed limit $v_{\text{limit}}$ or falling below the time-to-collision threshold $t_{\text{TTC}}$ incurs a fixed per-step penalty of $-0.1$. The time to collision is computed w.r.t.\ the lead vehicle as the longitudinal gap divided by the closing speed. The per-step reward is the sum of all rows in Tab.~\ref{tab:reward-conditioning}; the goal reward uses a fixed radius $\delta_{\text{goal}} = 2\,$m and is not part of the conditioning vector. The reward coefficients that form the behavior of our three traffic agents are presented in Tab.~\ref{tab:traffic-profiles}.

The behavior of the conditioned traffic agents is controlled by the reward coefficients, which are exposed as plain configuration entries, so new driving styles can be defined without modifying the training or evaluation pipeline. To examine whether each coefficient affects the behavior it is intended to control, we change one parameter of the $\color{purple}\pi_\theta^{\text{norm}}$ profile at a time and evaluate the resulting policy against IDM traffic on Interactive1k (Tab.~\ref{tab:coeff_sensitivity}). Most parameters primarily affect their target metric. Halving the speed limit reduces the mean speed by $2.23$\,m/s, and lowering the collision penalty by an order of magnitude increases the collision rate from $3.7\%$ to $4.9\%$. Disabling the TTC term ($t_{\text{TTC}}\!:3\!\to\!0$) has a similar effect on the collision rate ($+1.1$ percentage points) and additionally increases the mean speed by $0.74$\,m/s, i.e., the agent keeps smaller safety margins. Removing the comfort term increases comfort violations from $0.62$ to $5.12$ per episode, and removing the centerline term increases the distance to the centerline by $0.047$\,m. Side effects on the remaining metrics are small, with two exceptions that reflect couplings between objectives. Removing the comfort penalty also increases the mean speed ($+0.84$\,m/s), and a lower speed limit increases comfort violations ($+4.21$ per episode), presumably because the agent decelerates more often to stay within the limit. Only the alignment coefficient shows no clear effect: reducing it by almost two orders of magnitude leaves the lane-misalignment fraction essentially unchanged ($-0.001$). Based on our observations, lane alignment is largely determined by the remaining reward terms rather than by the alignment coefficient itself. Overall, the coefficients provide largely independent control over the individual behavioral dimensions of the traffic agents.

\begin{table*}[h]
\centering
\setlength{\aboverulesep}{0pt}
\setlength{\belowrulesep}{0pt}
\renewcommand{\arraystretch}{1.25}
\resizebox{\textwidth}{!}{%
\begin{tabular}{l|c|cccccc}
\toprule
Metric & $\color{purple}\pi_\theta^{\text{norm}}$ &
\makecell{$v_{\text{limit}}$\\$20\!\to\!10$} &
\makecell{$\alpha_{\text{collision}}$\\$3\!\to\!0.3$} &
\makecell{$\alpha_{\text{comfort}}$\\$0.015\!\to\!0$} &
\makecell{$\alpha_{\text{align}}$\\$1.3{\times}10^{-2}\!\to\!2.5{\times}10^{-4}$} &
\makecell{$\alpha_{\text{center}}$\\$0.02\!\to\!0$} &
\makecell{$t_{\text{TTC}}$\\$3\!\to\!0$} \\
\midrule
At-Fault Collision rate (\%)      & $3.7$   & \dltm{0.1}   & \underline{\dltp{1.2}} & \dltp{0.2}   & \dltp{0.3}   & \dltp{0.2}   & \underline{\dltp{1.1}} \\
Offroad rate (\%)        & $1.3$   & \dltm{0.6}   & \dltp{0.3}   & \dltp{0.3}   & \dltm{0.4}   & \dltm{0.2}   & \dltm{0.3} \\
Comfort violations / ep. & $0.62$  & \dltp{4.21}  & \dltm{0.15}  & \underline{\dltp{4.50}} & \dltp{0.02} & \dltm{0.03} & \dltm{0.39} \\
Centerline dist. (m)     & $0.421$ & \dltm{0.030} & \dltp{0.013} & \dltp{0.016} & \dltm{0.004} & \underline{\dltp{0.047}} & \dltz \\
Lane misalign. frac.     & $0.100$ & \dltm{0.001} & \dltm{0.002} & \dltp{0.003} & \underline{\dltm{0.001}} & \dltp{0.006} & \dltm{0.003} \\
Mean speed (m/s)         & $8.95$  & \underline{\dltm{2.23}} & \dltp{0.21} & \dltp{0.84} & \dltm{0.02} & \dltp{0.01} & \dltp{0.74} \\
\bottomrule
\end{tabular}%
}
\caption{Parameter sensitivity of the conditioned ego policy against IDM traffic
on Interactive1k. Each column changes a single parameter of the
$\color{purple}\pi_\theta^{\text{norm}}$ profile, keeping all others fixed.
The $\color{purple}\pi_\theta^{\text{norm}}$ column reports the mean over all
scenarios; the remaining columns report the deviation from it
(\dltp{} increase, \dltm{} decrease, \dltz{} no change).
\underline{Underlined} entries mark the metric each change is expected to
affect primarily.}
\label{tab:coeff_sensitivity}
\end{table*}

\begin{table}[h]
\centering
\renewcommand{\arraystretch}{1.6}
\resizebox{\textwidth}{!}{%
\begin{tabular}{@{}ll@{}}
\toprule
\textbf{Reward} & \textbf{Training distribution} \\
\midrule
$R_{\text{goal}} = \mathbb{I}_{\Vert x - g \Vert < \delta_{\text{goal}}}$
  & $\delta_{\text{goal}} = 2\,\text{m}$ \\
\addlinespace[2pt]\midrule
$R_{\text{collision}} = -(\alpha_{\text{collision}} + 0.1\,|v|)\,\mathbb{I}_{\text{collision}}$
  & $\alpha_{\text{collision}} \sim \mathcal{U}(0, 3)$ \\
\addlinespace[2pt]\midrule
$R_{\text{off-road}} = -\alpha_{\text{boundary}}\,\mathbb{I}_{\text{boundary}}$
  & $\alpha_{\text{boundary}} \sim \mathcal{U}(0, 3)$ \\
\addlinespace[2pt]\midrule
$R_{\text{speed}} = -0.1\,\mathbb{I}_{|v| > v_{\text{limit}}}$
  & $v_{\text{limit}} \sim \mathcal{U}(5, 50)$ \\
\addlinespace[2pt]\midrule
$R_{\text{TTC}} = -0.1\,\mathbb{I}_{\text{TTC} < t_{\text{TTC}}}$
  & $t_{\text{TTC}} \sim \mathcal{U}(0, 10)$ \\
\addlinespace[2pt]\midrule
$R_{\text{comfort}} = -\alpha_{\text{comfort}}\,\Bigl(
   \mathbb{I}_{|a_{\text{long}}|>3} + \mathbb{I}_{|a_{\text{lat}}|>3}
   + \mathbb{I}_{|\dot a_{\text{long}}|>5 \,\vee\, |\dot a_{\text{lat}}|>5}\Bigr)$
  & $\alpha_{\text{comfort}} \sim \mathcal{U}(0, 0.0333)$ \\
\addlinespace[2pt]\midrule
$R_{\text{l-align}} = \alpha_{\text{l-align}}\,\Delta t \Bigl(
   \min(\cos\theta_f, 0)
   + \alpha_{\text{vel-align}} \min(\cos\theta_f \cdot v, 0)
   + 0.0025\,(1 - \tfrac{|\theta_f|}{\pi/2}) \Bigr)$
  & \makecell[l]{$\alpha_{\text{l-align}} \sim \mathcal{U}(2.5 \times 10^{-4}, 2.5 \times 10^{-2})$ \\
                 $\alpha_{\text{vel-align}} \sim \mathcal{U}(0, 1)$} \\
\addlinespace[2pt]\midrule
$R_{\text{l-center}} = -\alpha_{\text{l-center}}\,\Delta t \Bigl(
   \mathbb{I}_{\cos\theta_f > 0.5} \cdot |x_f - \alpha_{\text{center-bias}}|
   - \dfrac{0.05}{\exp(|x_f - \alpha_{\text{center-bias}}| - 0.5)} \Bigr)$
  & \makecell[l]{$\alpha_{\text{l-center}} \sim \mathcal{U}(0, 5 \times 10^{-2})$ \\
                 $\alpha_{\text{center-bias}} \sim \mathcal{U}(-0.5, 0.5)$} \\
\addlinespace[2pt]\midrule
$R_{\text{velocity}} = \alpha_{\text{velocity}}\,\Delta t \cdot \max(\cos\theta_f, 0)\,\mathbb{I}_{|v|>2.5}$
  & $\alpha_{\text{velocity}} = 2.5 \times 10^{-3}$ \\
\addlinespace[2pt]\midrule
$R_{\text{reverse}} = -\alpha_{\text{reverse}}\,\Delta t\,\mathbb{I}_{v < 0}$
  & $\alpha_{\text{reverse}} \sim \mathcal{U}(2.5 \times 10^{-4}, 7.5 \times 10^{-3})$ \\
\addlinespace[2pt]\midrule
$R_{\text{timestep}} = -\alpha_{\text{timestep}}\,\Delta t \cdot \mathbb{I}_{|v|>0 \,\vee\, |a|>0}$
  & $\alpha_{\text{timestep}} = 2.5 \times 10^{-5}$ \\
\bottomrule
\end{tabular}%
}
\caption{Reward components and their training-time sampling distributions. Terms with $\sim$ are per-agent, per-episode sampled conditioning coefficients, fixed values are identical for all agents and not part of the conditioning vector. $v_{\text{limit}}$ in m/s, $t_{\text{TTC}}$ in s.}
\label{tab:reward-conditioning}
\end{table}

\section{Traffic Agents}
\label{sec:traffic-agents}
The coefficients of the three conditioned traffic agents can be found in Tab.~\ref{tab:traffic-profiles}. We use similar IDM parameters as in nuPlan and provide them in Tab.~\ref{tab:idm-params}.
\begin{table}[H]
\centering
\begin{minipage}[H]{0.55\linewidth}
    \centering
    \small
    \begin{tabular}{@{}lccc@{}}
    \toprule
    \textbf{Coefficient} & {\color{red}\textbf{Aggressive}} & {\color{purple}\textbf{Normal}} & {\color{green}\textbf{Cautious}} \\
                         & {\color{red}$\pi_\theta^{\text{aggr}}$} & {\color{purple}$\pi_\theta^{\text{norm}}$} & {\color{green}$\pi_\theta^{\text{caut}}$} \\
    \midrule
    $v_{\text{limit}}$             & {\color{red}$45$}                 & {\color{purple}$20$}                 & {\color{green}$10$}                \\
    $\alpha_{\text{collision}}$    & {\color{red}$0.3$}                & {\color{purple}$3.0$}                & {\color{green}$3.0$}               \\
    $\alpha_{\text{boundary}}$     & {\color{red}$0.3$}                & {\color{purple}$3.0$}                & {\color{green}$3.0$}               \\
    $\alpha_{\text{comfort}}$      & {\color{red}$0.0$}                & {\color{purple}$1.5 \times 10^{-2}$} & {\color{green}$3.33 \times 10^{-2}$} \\
    $\alpha_{\text{l-align}}$      & {\color{red}$1.0 \times 10^{-3}$} & {\color{purple}$1.3 \times 10^{-2}$} & {\color{green}$2.5 \times 10^{-2}$} \\
    $\alpha_{\text{vel-align}}$    & {\color{red}$0.1$}                & {\color{purple}$0.5$}                & {\color{green}$1.0$}               \\
    $\alpha_{\text{l-center}}$     & {\color{red}$2.0 \times 10^{-3}$} & {\color{purple}$2.0 \times 10^{-2}$} & {\color{green}$5.0 \times 10^{-2}$} \\
    $\alpha_{\text{center-bias}}$  & {\color{red}$0.0$}                & {\color{purple}$0.0$}                & {\color{green}$0.0$}               \\
    $\alpha_{\text{reverse}}$      & {\color{red}$5.0 \times 10^{-4}$} & {\color{purple}$4.0 \times 10^{-3}$} & {\color{green}$7.5 \times 10^{-3}$} \\
    $t_{\text{TTC}}$               & {\color{red}$1$}                  & {\color{purple}$3$}                  & {\color{green}$8$}                 \\
    \bottomrule
    \end{tabular}
    \caption{Reward coefficients defining the three conditioned traffic behaviors. $v_{\text{limit}}$ in m/s, $t_{\text{TTC}}$ in s.}
    \label{tab:traffic-profiles}
\end{minipage}%
\hfill
\begin{minipage}[H]{0.42\linewidth}
    \centering
    \small
    \begin{tabular}{lccl}
    \toprule
    \textbf{Parameter} & \textbf{Symbol} & \textbf{Value} & \textbf{Unit} \\
    \midrule
    Target velocity       & $v_0$    & 15.0 & m/s \\
    Minimum gap           & $s_0$    & 1.0  & m \\
    Desired time headway  & $T$      & 1.5  & s \\
    Maximum acceleration  & $a$      & 1.0  & m/s\textsuperscript{2} \\
    Maximum deceleration  & $b$      & 3.0  & m/s\textsuperscript{2} \\
    Acceleration exponent & $\delta$ & 4.0  & -- \\
    \bottomrule
    \end{tabular}
    \caption{Intelligent Driver Model (IDM) parameters.}
    \label{tab:idm-params}
\end{minipage}
\end{table}

\section{Runtime}
\label{sec:runtime}
We compare the wall-clock time of closed-loop simulation in BehaviorBench with that of the nuPlan simulator~\citep{nuplan} in Tab.~\ref{tab:timing}. To isolate the simulator overhead, both simulators run the same IDM  planner against reactive IDM traffic on the same 100 scenarios, each rolled out for 149 steps (15\,s at 10\,Hz). BehaviorBench simulates a scenario in $1.84$\,s on average, compared to $27.25$\,s for nuPlan, a $14.8\times$ speedup.
\begin{table}[h]
\centering
\small
\begin{tabular}{lc}
\toprule
Simulator & Time / scenario (s) \\
\midrule
nuPlan (closed-loop, reactive IDM)  & $27.25 \pm 14.93$ \\
BehaviorBench  (IDM)                & $\phantom{0}1.84 \pm \phantom{0}0.85$ \\
\midrule
Speedup & $14.8\times$ \\
\bottomrule
\end{tabular}
\caption{Simulation wall-clock time on the same 100 scenarios, run single-threaded on one core of an AMD EPYC 9334 with IDM ego and reactive IDM traffic. Each scenario is simulated for a fixed 149 steps (15\,s at 10\,Hz) in both simulators. Values are mean\,$\pm$\,std over the 100 scenarios.}
\label{tab:timing}
\end{table}
\section{Hybrid Planner: PDM+PPO}
\label{sec:app_pdm_ppo}
The motivation behind our approach is to combine the safety of rule-based planners such as PDM with the assertiveness of a policy trained via PPO and a simple reward. Our experiments indicate that PDM performs better in routine driving scenarios, whereas a policy trained via PPO and a simple reward is more effective in interactive ones, suggesting that the two paradigms are complementary rather than competing. A second motivation stems from the well-known difficulty of designing reward functions for driving, which typically demands extensive tuning and the careful integration of many driving principles into a single complex objective. Instead, we train a policy with only a small set of high-level driving principles, namely off-road avoidance, collision avoidance, and goal completion. By rolling out multiple high-probability actions from this policy and scoring the resulting trajectories with a comprehensive scoring function, we additionally enforce lower-level principles such as comfort, lane alignment, and centered driving at planning time rather than during reward engineering.

In our nuPlan evaluation, the comparatively slow simulation prevented us from performing rollouts at inference time, so we instead selected the most likely action from the policy. This highlights a key advantage of PufferDrive: its high simulation throughput makes it feasible to use the simulator itself as the forward model and to evaluate multiple rollouts within a reasonable planning budget, which is not feasible in nuPlan.
For PDM we used a horizon of 4\,s, while for the PPO fallback we selected the top-8 actions from the policy $\pi_\theta$ and rolled each out for 1\,s. Pseudocode for the hybrid planner is provided in Alg.~\ref{alg:hybrid_pdm_ppo}.

\begin{figure}[ht]
    \includegraphics[width=\textwidth]{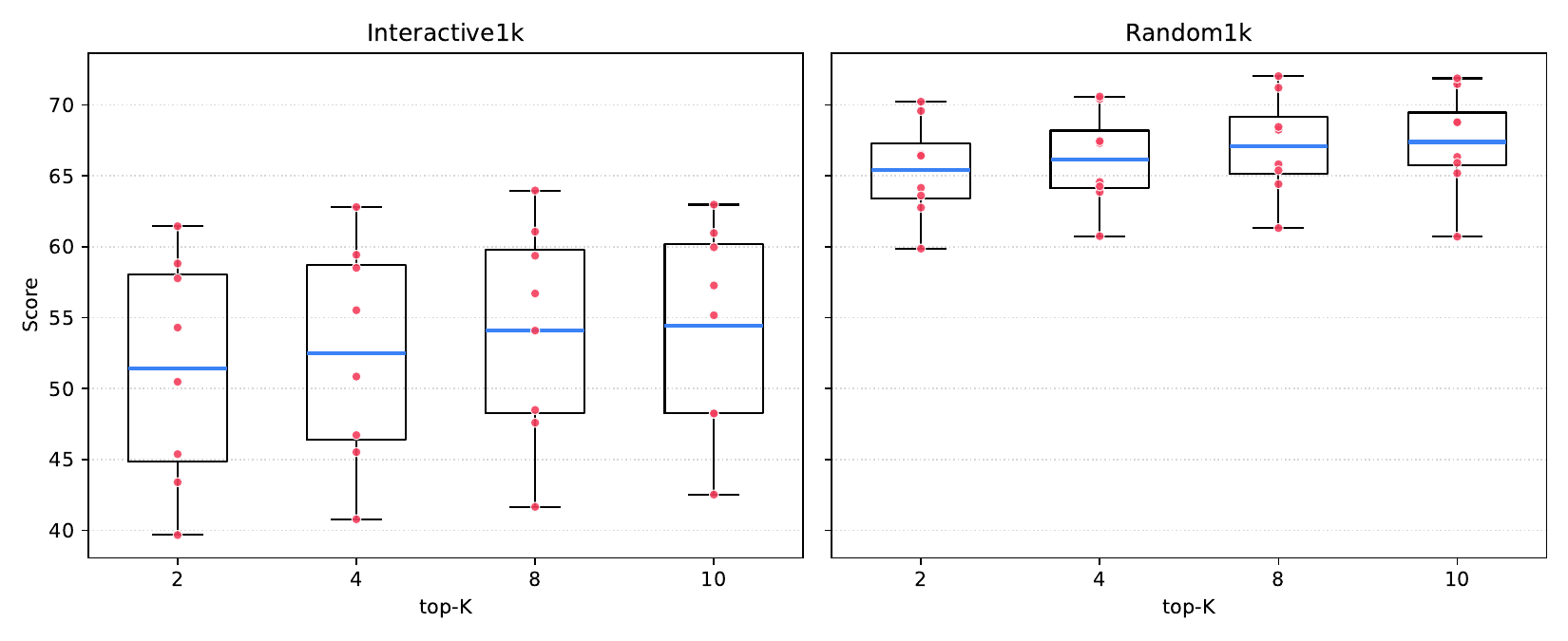}
    \caption{Score of the PDM+PPO hybrid planner as a function of top-K on Interactive1k (left) and Random1k (right). Each red marker reports the mean score (scaled by 100) over all scenarios for one traffic agent type, and the box summarizes the resulting distribution across traffic agent types at a fixed planning horizon of 10 steps (1\,s). Performance improves with increasing top-K on both splits, with diminishing returns beyond $K = 8$. The spread across traffic conditions is markedly larger on Interactive1k than on Random1k, reflecting the heavier reactive demands of the interactive scenarios.}
    \vspace{-0.4cm}
    \label{fig:pdm_ppo_topk_performance}
\end{figure}

\begin{figure}[ht]
    \includegraphics[width=\textwidth]{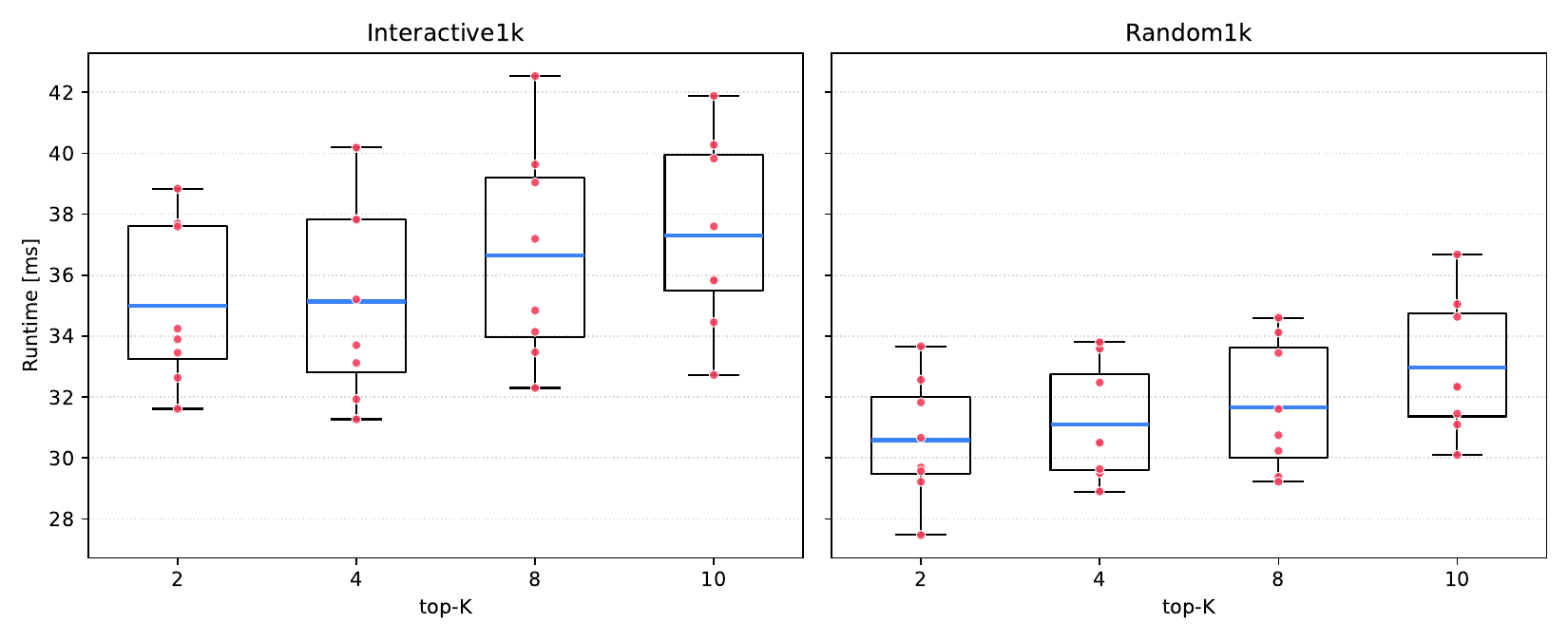}
    \caption{Per-step runtime of the PDM+PPO hybrid planner as a function of top-K on Interactive1k (left) and Random1k (right). Each red marker reports the mean per-step runtime in ms over all scenarios for one traffic agent type, and the box summarizes the distribution across traffic agent types at a fixed planning horizon of 10 steps (1\,s). Runtime grows with top-K on both splits and remains consistently higher on Interactive1k than on Random1k, reflecting the additional cost induced by denser scenarios.}
    \vspace{-0.4cm}
    \label{fig:pdm_ppo_topk_runtime}
\end{figure}

We ablate the choice of $K$ in Fig.~\ref{fig:pdm_ppo_topk_performance} and observe that performance improves monotonically with $K$, with diminishing returns beyond $K=8$.
Fig.~\ref{fig:pdm_ppo_topk_runtime} reports the corresponding runtime analysis of our hybrid planner. Per-step runtime grows with $K$ but remains well within real-time budgets across all configurations. Compared to the original PDM implementation, we reimplement the IDM simulation in C rather than Python, yielding a substantial speed-up.

\begin{figure}[ht]
    \includegraphics[width=\textwidth]{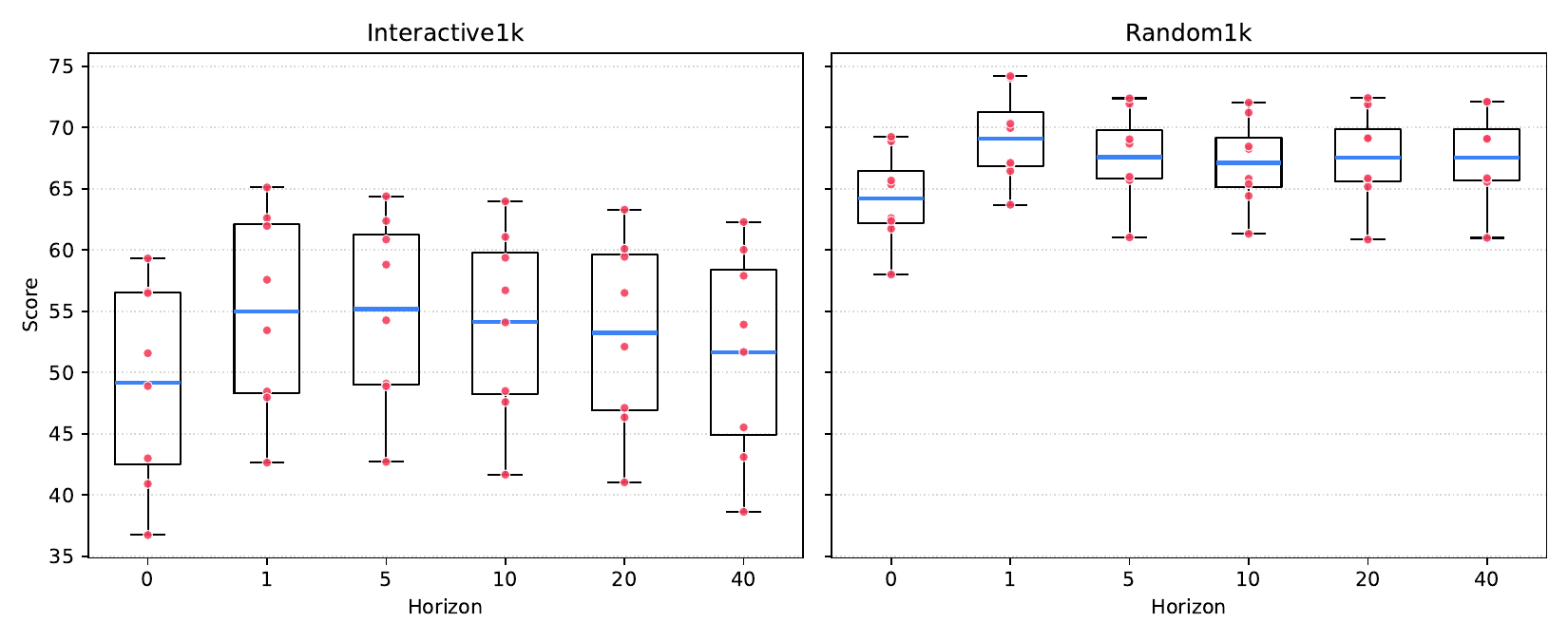}
    \caption{Score of the PDM+PPO hybrid planner as a function of the planning horizon $H_2$ on Interactive1k (left) and Random1k (right). Each red marker reports the mean score over all scenarios for one traffic agent type, and the box summarizes the distribution across traffic agent types at a fixed top-K with $K=8$.}
    \vspace{-0.4cm}
    \label{fig:pdm_ppo_horizon_performance}
\end{figure}

\begin{figure}[ht]
    \includegraphics[width=\textwidth]{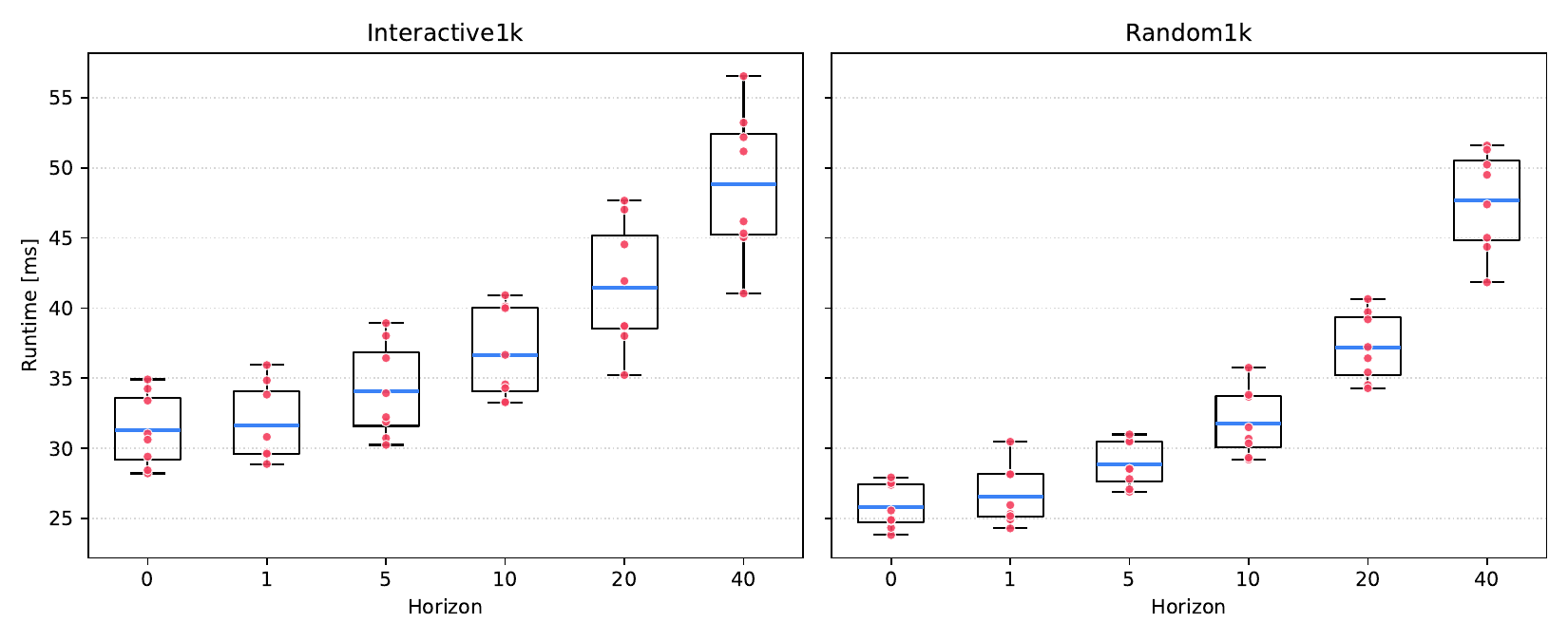}
    \caption{Per-step runtime of the PDM+PPO hybrid planner as a function of the planning horizon on Interactive1k (left) and Random1k (right). Each red marker reports the mean per-step runtime in ms over all scenarios for one traffic agent type, and the box summarizes the distribution across traffic agent types at a fixed top-K with $K=8$. Runtime grows with the horizon on both splits and remains consistently higher on Interactive1k than on Random1k, reflecting the additional cost induced by denser scenarios.}
    \vspace{-0.4cm}
    \label{fig:pdm_ppo_horizon_runtime}
\end{figure}

We further ablate the planning horizon $H_2$ in Fig.~\ref{fig:pdm_ppo_horizon_performance} and observe a sharp jump in score from $H_2=0$ to $H_2=1$ on both splits. On Interactive1k, performance peaks at short horizons and gradually declines beyond $H_2=10$, which we attribute to the fixed agent predictions used during rollout, extending the horizon propagates the assumption that surrounding agents follow predetermined trajectories unaffected by the ego vehicle, biasing trajectory selection towards overly conservative behavior in densely interactive scenes. On Random1k, by contrast, the score saturates after $H_2=5$ as these scenarios contain less interaction. Fig.~\ref{fig:pdm_ppo_horizon_runtime} shows that per-step runtime grows monotonically with $H_2$, but even at the largest evaluated horizon of $H_2=40$ ($4\,$s) the runtime stays well below the $100\,$ms budget required for real-time operation at $10\,$Hz.
\begin{algorithm}[t]
\caption{Our Hybrid PDM+PPO Planner}
\label{alg:hybrid_pdm_ppo}
\begin{algorithmic}[1]
\Require state $s_t$, policy $\pi_\theta$, rollout horizons $H_1$, $H_2$, number of rollouts $K$
\State $Y \gets \textsc{CVPrediction}(s_t, H_1)$ \Comment{constant-velocity forecast of surrounding agents}
\State $\{\tau_i, c_i\}_{i=1}^{15} \gets \textsc{PDM}(s_t, Y)$ \Comment{IDM proposals scored by PDM}
\If{$\max_i c_i > 0$}    \State $i^\star = \arg\max_i c_i$
    \State \Return $\tau_{i^\star}$
\EndIf
\State $\{a_k\}_{k=1}^{K} \gets \textsc{TopK}\bigl(\pi_\theta(s_t)\bigr)$ \Comment{fallback: top-$K$ policy $\pi_\theta$ actions}
\For{$k = 1, \dots, K$}
    \State $\tau_k \gets \textsc{Rollout}(s_t, a_k, Y)$ \Comment{$H_2$-step rollout under forecast $Y$}
    \State $c_k \gets \textsc{Score}(\tau_k)$ \Comment{collision, off-road, progress, comfort, alignment, centerline}
\EndFor
\State $k^\star = \arg\max_k c_k$ 
\State \Return $\tau_{k^\star} $
\end{algorithmic}
\end{algorithm}

\clearpage
\section{WOSAC Realism Evaluation}
\label{sec:wosac-eval}

We evaluate driving realism using the Waymo Open Sim Agents Challenge (WOSAC) 2024 metrics~\citep{montali2024wosac}.
All agents in a scenario are controlled by the same planner policy.
For each scenario, we perform $K=32$ stochastic rollouts starting from a shared initial state observed from the log for $T_\text{init}=10$ steps.
The simulation then runs for $T_\text{sim}=81$ steps at $\Delta t = 0.1\,\text{s}$ (total episode length 91 steps $= 9.1\,\text{s}$). An overview of the parameters can be found in Tab.~\ref{tab:wosac-params}.

\begin{table}[ht]
\centering
\caption{WOSAC evaluation parameters.}
\label{tab:wosac-params}
\begin{tabular}{lrl}
\toprule
\textbf{Parameter} & \textbf{Value} & \textbf{Description} \\
\midrule
Episode length     & 91 steps    & Total scenario duration ($9.1\,\text{s}$) \\
$\Delta t$         & $0.1\,\text{s}$ & Simulation timestep \\
Init steps $T_\text{init}$ & 10  & GT-driven warm-up steps \\
Sim steps $T_\text{sim}$   & 81  & Policy-controlled steps \\
Num.\ rollouts $K$ & 32         & Stochastic rollouts per scenario \\
\bottomrule
\end{tabular}
\end{table}

\section{Additional Results}
\label{sec:additional_res}
We provide the per-subscore results of our evaluation on Interactive1k and Random1k in Tab.~\ref{tab:planner_vs_traffic_reactive} and Tab.~\ref{tab:planner_vs_traffic_behavior}, as well as the per-subscore WOSAC results for our conditioned traffic agents in Tab.~\ref{tab:realism_subscores}.

\begin{figure}[ht]
    \includegraphics[width=\textwidth]{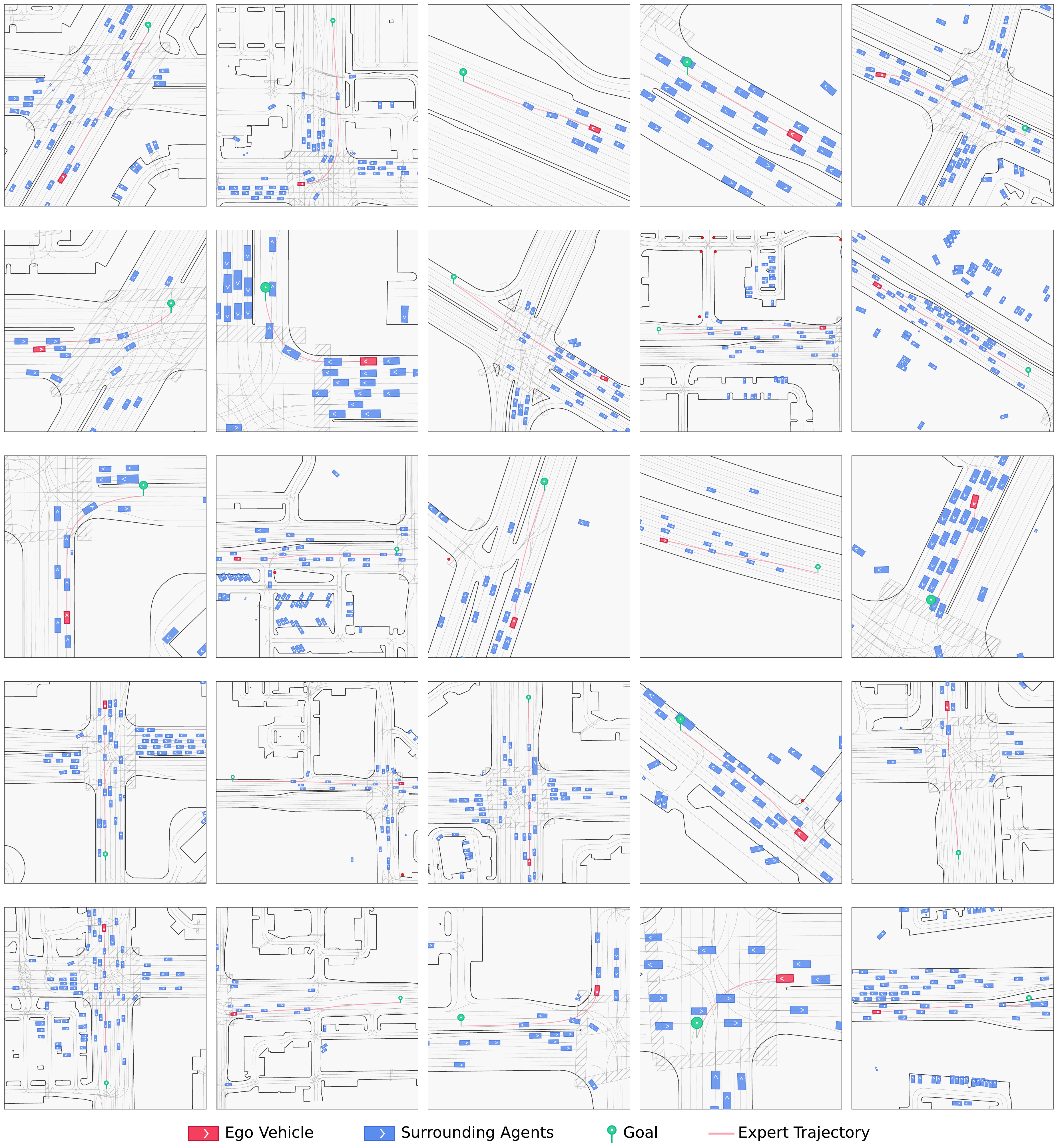}
    \caption{Example scenarios from our Interactive1k split.}
    \vspace{-0.4cm}
    \label{fig:app_interactive1k}
\end{figure}

\begin{figure}[ht]
    \includegraphics[width=\textwidth]{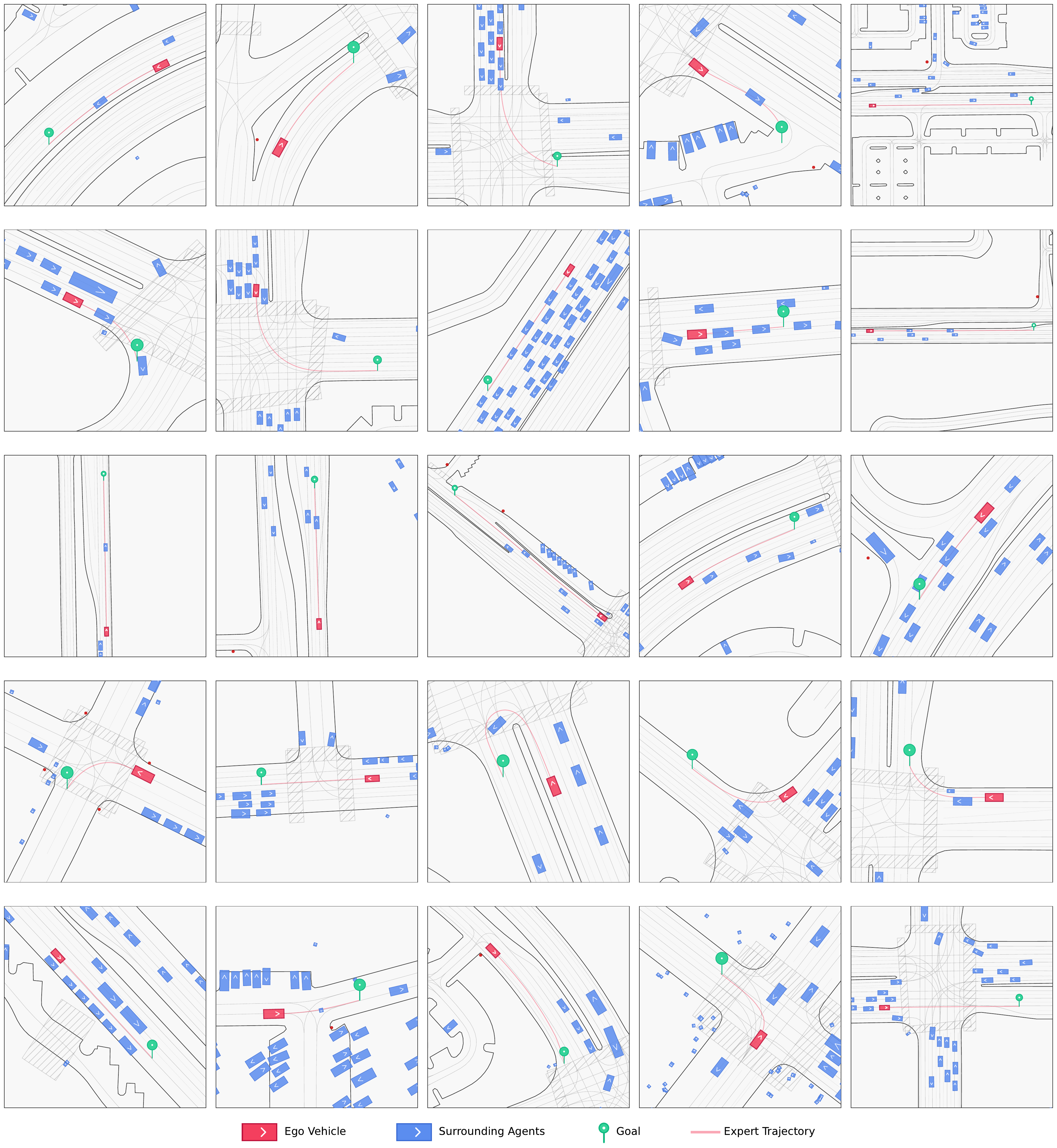}
    \caption{Example scenarios from our Random1k split.}
    \vspace{-0.4cm}
    \label{fig:app_random1k}
\end{figure}

Additional qualitative results for Interactive1k and Random1k are shown in Fig.~\ref{fig:app_interactive1k} and Fig.~\ref{fig:app_random1k}. These examples confirm that Interactive1k contains denser and more interactive scenarios, often requiring lane changes and complex interactions, whereas Random1k consists of simpler scenarios with less interaction.

\begin{figure}[t]
    \centering
    \includegraphics[width=\linewidth]{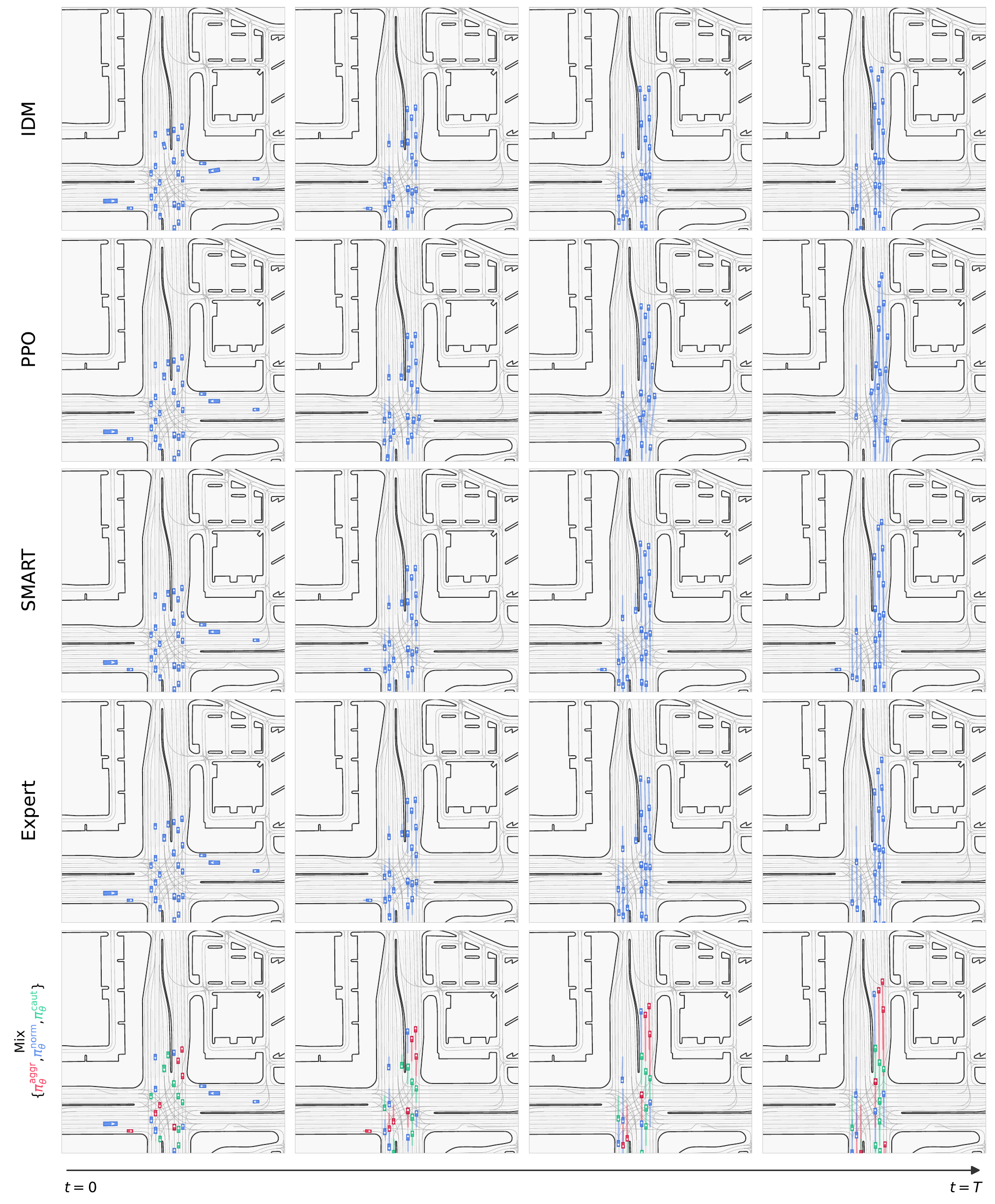}
    \caption{
        \textbf{Qualitative comparison of traffic agent models.}
        We visualize rollouts of IDM, PPO, SMART, expert and the conditioned PPO traffic agents (top to bottom) on the same scenario at four time steps over $T=3\,\mathrm{s}$ (left to right). Agents are shown with their past trajectories.
    }
    \vspace{-0.4cm}
    \label{fig:qualitative_traffic_agents_1}
\end{figure}

\begin{figure}[t]
    \centering
    \includegraphics[width=\linewidth]{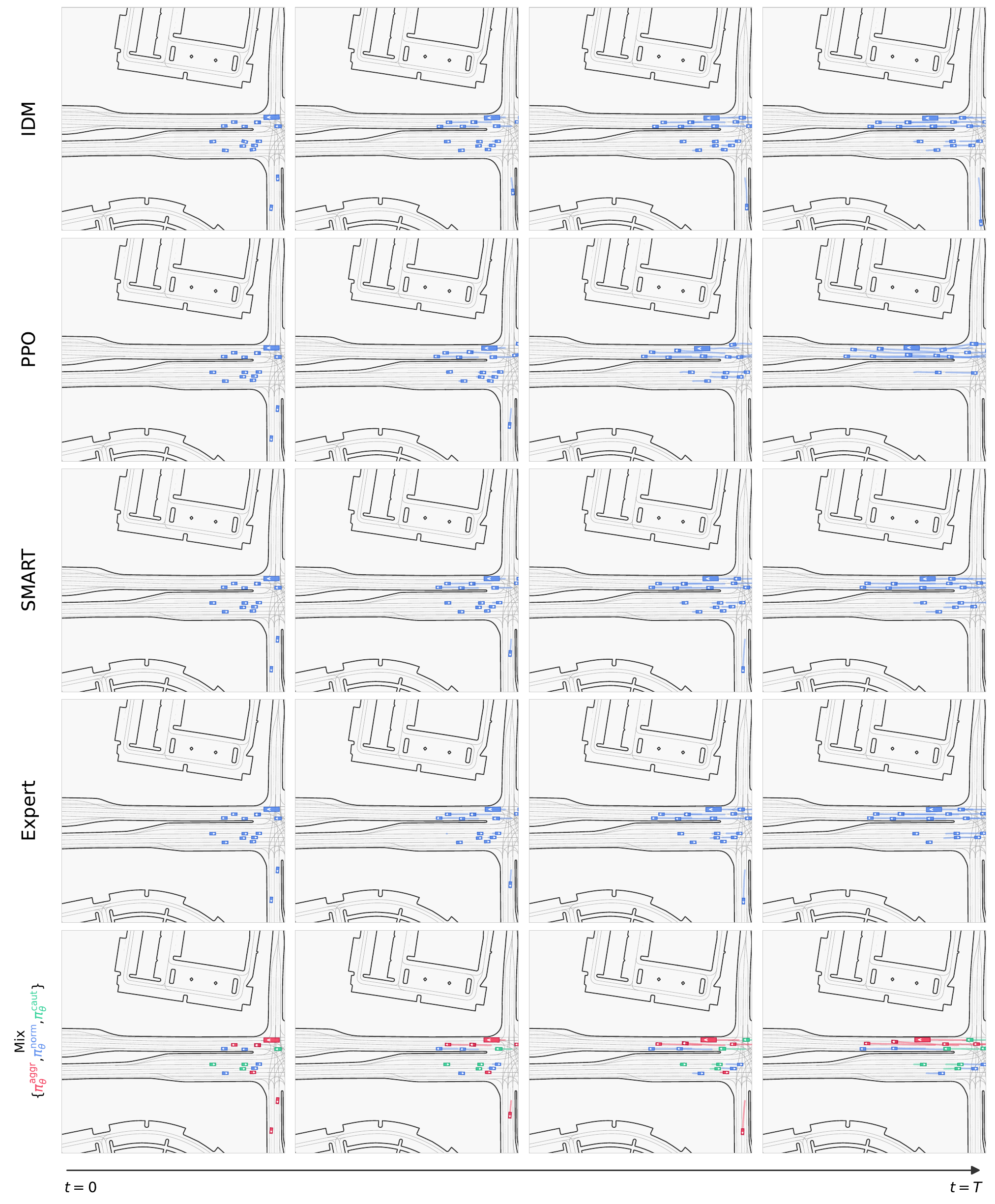}
    \caption{
        \textbf{Qualitative comparison of traffic agent models.}
        We visualize rollouts of IDM, PPO, SMART, expert and the conditioned PPO traffic agents (top to bottom) on the same scenario at four time steps over $T=3\,\mathrm{s}$ (left to right). Agents are shown with their past trajectories.
    }
    \vspace{-0.4cm}
    \label{fig:qualitative_traffic_agents_2}
\end{figure}

\begin{figure}[t]
    \centering
    \includegraphics[width=\linewidth]{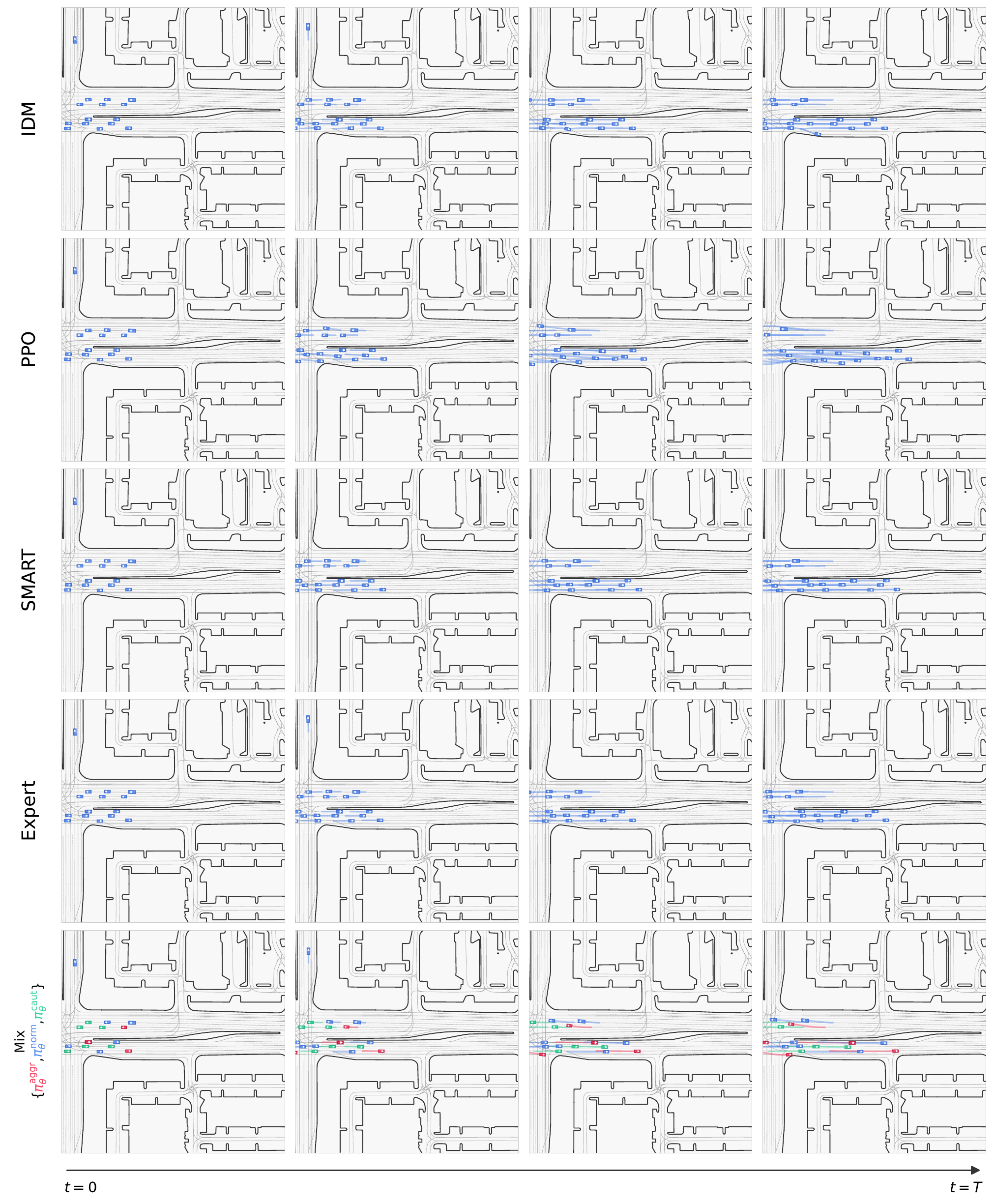}
    \caption{
        \textbf{Qualitative comparison of traffic agent models.}
        We visualize rollouts of IDM, PPO, SMART, expert and the conditioned PPO traffic agents (top to bottom) on the same scenario at four time steps over $T=3\,\mathrm{s}$ (left to right). Agents are shown with their past trajectories.
    }
    \vspace{-0.4cm}
    \label{fig:qualitative_traffic_agents_3}
\end{figure}

\begin{figure}[t]
    \centering
    \includegraphics[width=\linewidth]{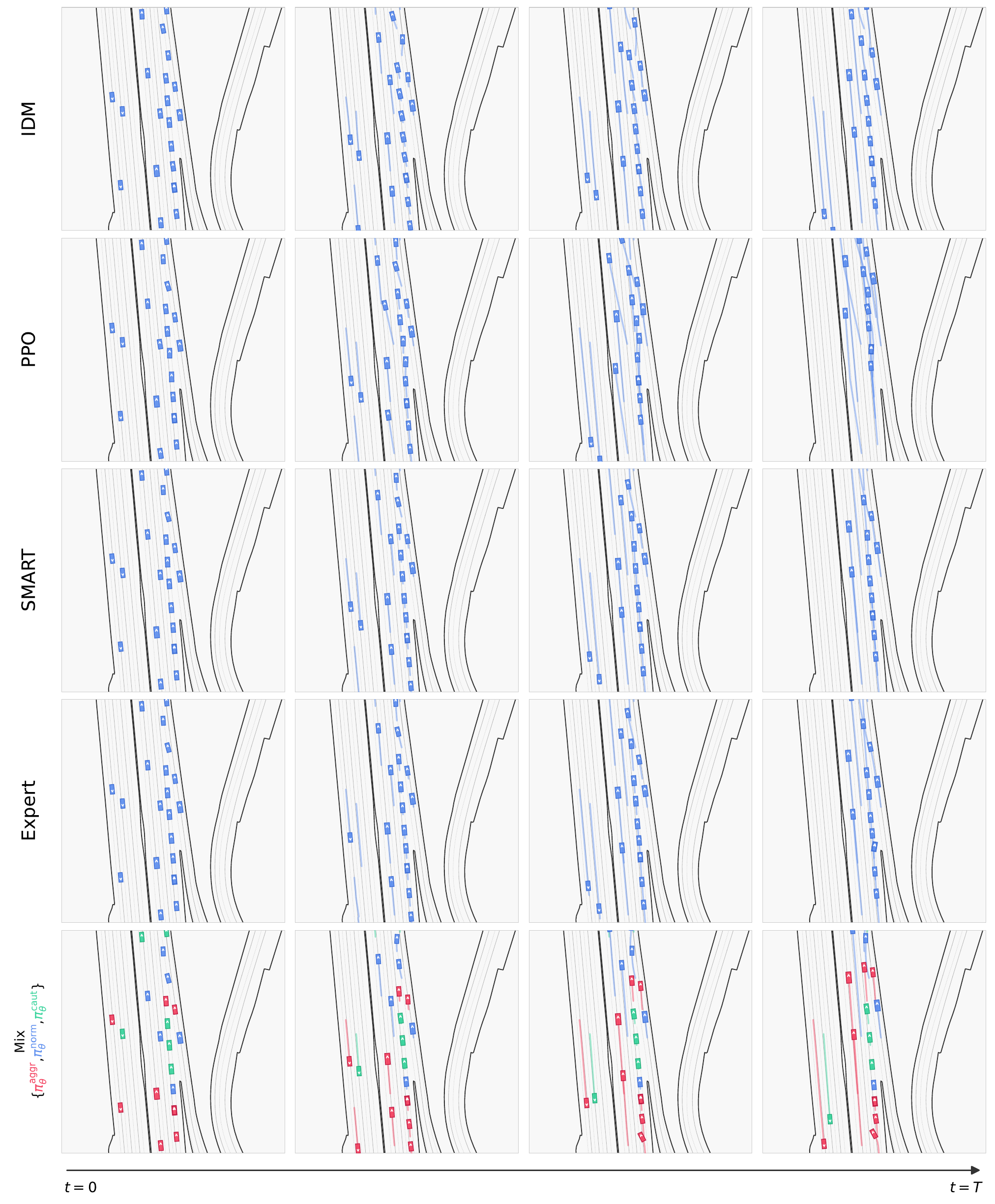}
    \caption{
        \textbf{Qualitative comparison of traffic agent models.}
        We visualize rollouts of IDM, PPO, SMART, expert and the conditioned PPO traffic agents (top to bottom) on the same scenario at four time steps over $T=3\,\mathrm{s}$ (left to right). Agents are shown with their past trajectories.
    }
    \vspace{-0.4cm}
    \label{fig:qualitative_traffic_agents_4}
\end{figure}

\begin{figure}[t]
    \centering
    \includegraphics[width=\linewidth]{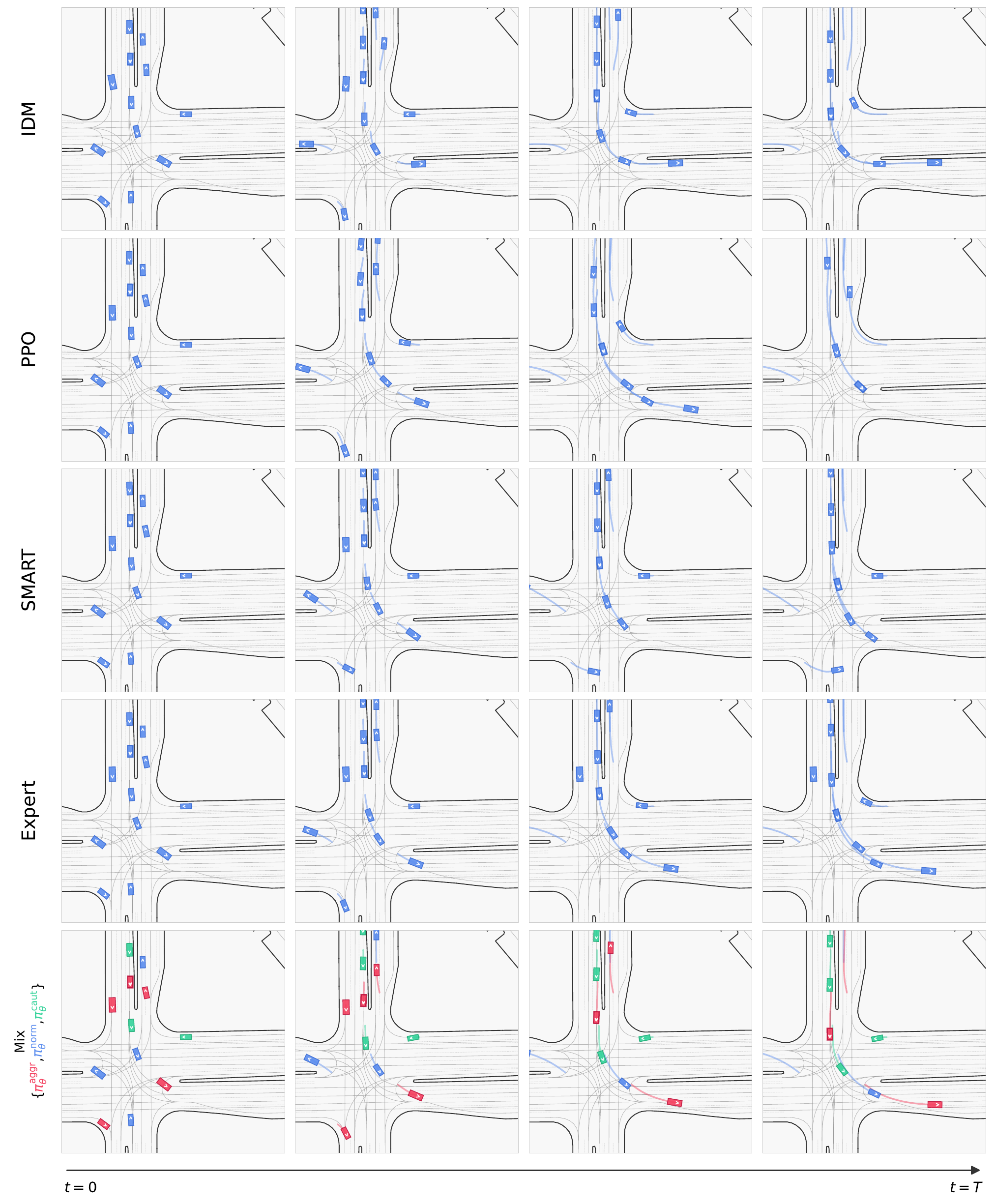}
    \caption{
        \textbf{Qualitative comparison of traffic agent models.}
        We visualize rollouts of IDM, PPO, SMART, expert and the conditioned PPO traffic agents (top to bottom) on the same scenario at four time steps over $T=3\,\mathrm{s}$ (left to right). Agents are shown with their past trajectories.
    }
    \vspace{-0.4cm}
    \label{fig:qualitative_traffic_agents_5}
\end{figure}

\begin{figure}[t]
    \centering
    \includegraphics[width=\linewidth]{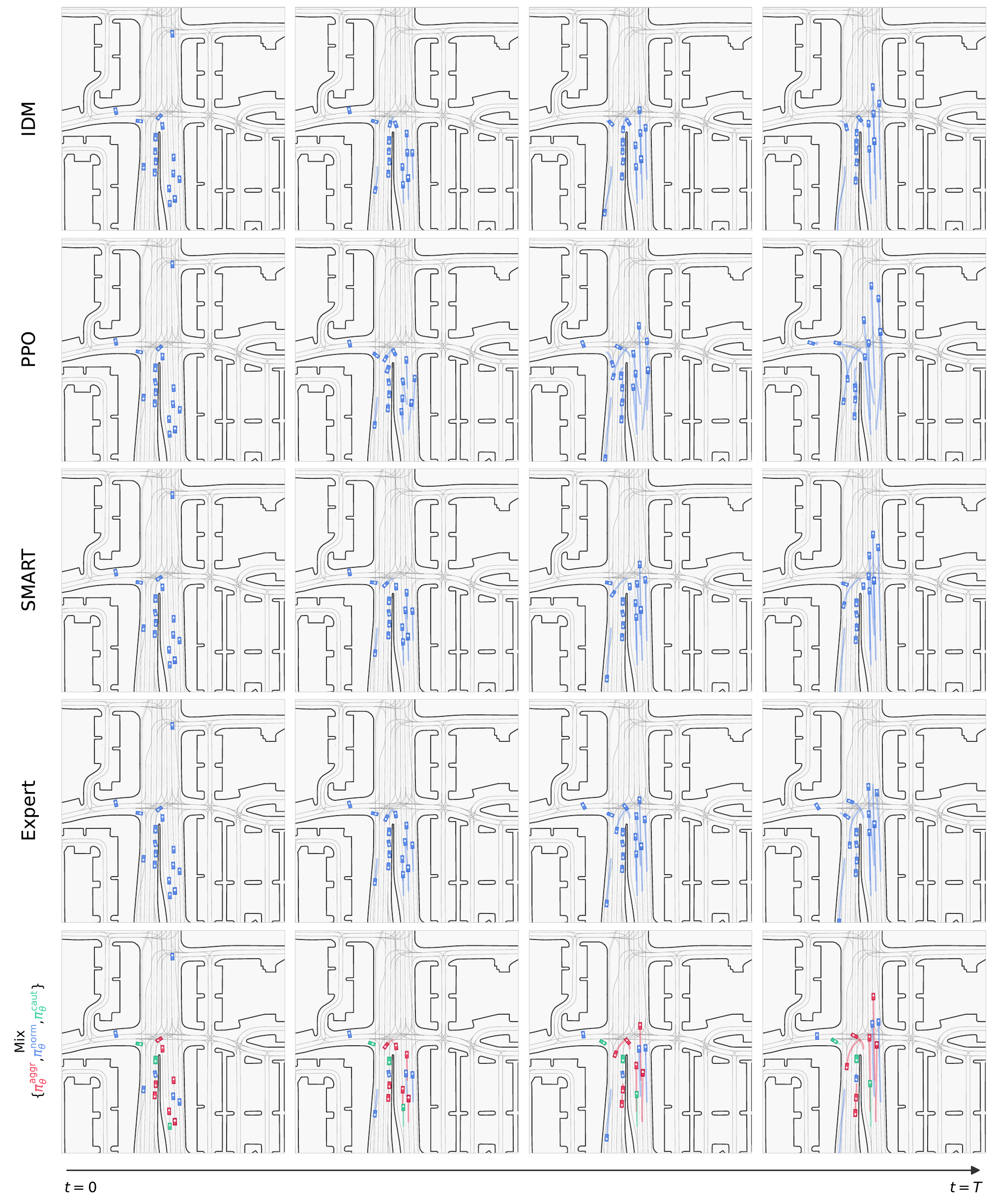}
    \caption{
        \textbf{Qualitative comparison of traffic agent models.}
        We visualize rollouts of IDM, PPO, SMART, expert and the conditioned PPO traffic agents (top to bottom) on the same scenario at four time steps over $T=3\,\mathrm{s}$ (left to right). Agents are shown with their past trajectories.
    }
    \vspace{-0.4cm}
    \label{fig:qualitative_traffic_agents_6}
\end{figure}

\begin{figure}[t]
    \centering
    \includegraphics[width=\linewidth]{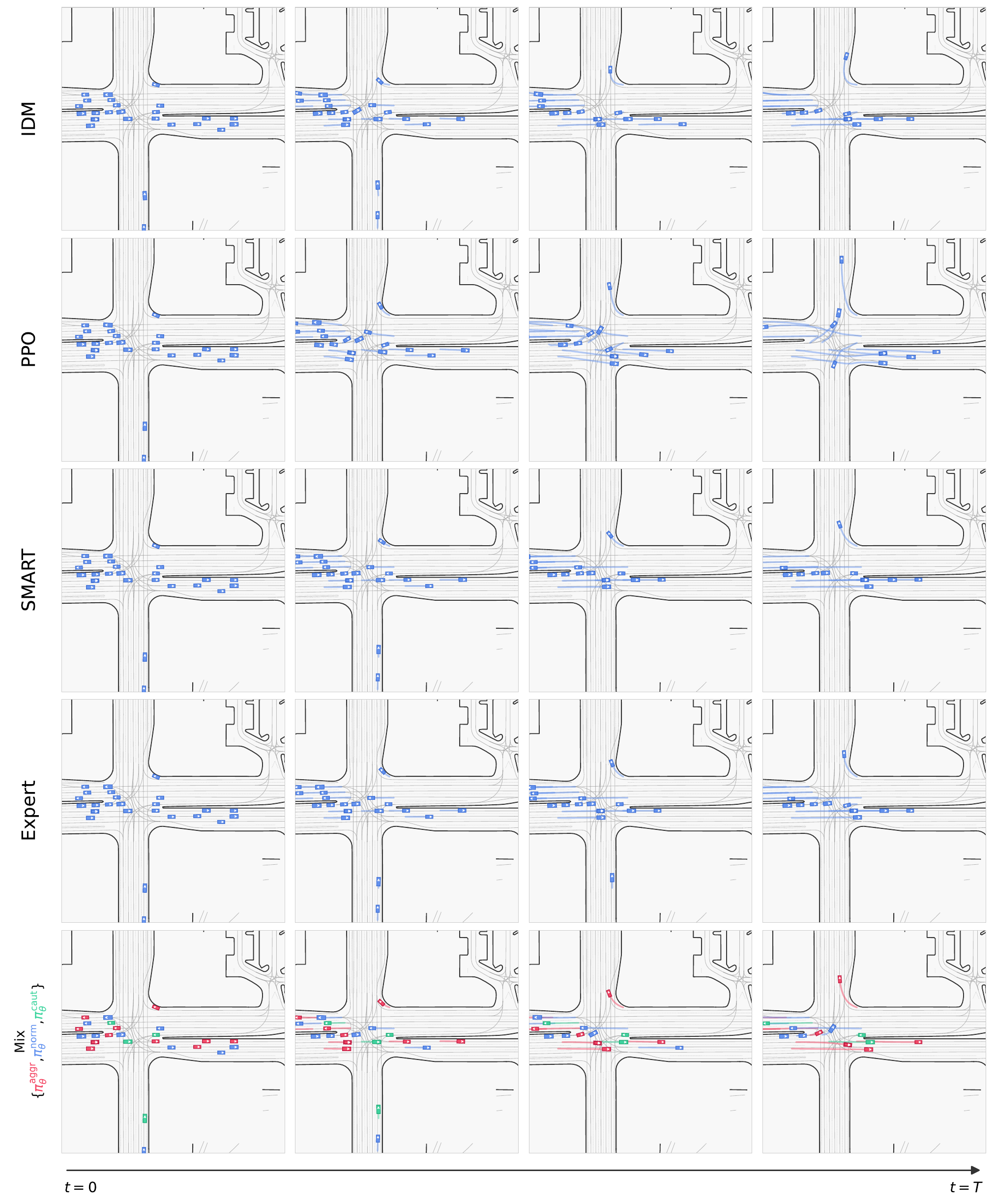}
    \caption{
        \textbf{Qualitative comparison of traffic agent models.}
        We visualize rollouts of IDM, PPO, SMART, expert and the conditioned PPO traffic agents (top to bottom) on the same scenario at four time steps over $T=3\,\mathrm{s}$ (left to right). Agents are shown with their past trajectories.
    }
    \vspace{-0.4cm}
    \label{fig:qualitative_traffic_agents_7}
\end{figure}

We also provide qualitative results of our traffic agents in \Cref{fig:qualitative_traffic_agents_1,fig:qualitative_traffic_agents_2,fig:qualitative_traffic_agents_3,fig:qualitative_traffic_agents_4,fig:qualitative_traffic_agents_5,fig:qualitative_traffic_agents_6,fig:qualitative_traffic_agents_7}.
We observe that the conditioned traffic agents exhibit behavior consistent with the intended conditioning. In particular, the aggressive agents in red drive faster than the other agent types and change lanes to overtake slower vehicles, as shown in Fig.~\ref{fig:qualitative_traffic_agents_2}, Fig.~\ref{fig:qualitative_traffic_agents_3}, and Fig.~\ref{fig:qualitative_traffic_agents_4}. The dense highway traffic in Fig.~\ref{fig:qualitative_traffic_agents_4} additionally highlights the different following behaviors: while an aggressive agent weaves across lane markings to get past the queue ahead, the aggressive agents closely tailgate each other, whereas the cautious agents in green keep noticeably larger gaps to their lead vehicles.
The cautious agents furthermore consistently yield at intersections. In Fig.~\ref{fig:qualitative_traffic_agents_7}, a cautious agent remains stopped at the entrance of the intersection for the entire horizon to let the crossing traffic pass, whereas the expert immediately enters and crosses it. In Fig.~\ref{fig:qualitative_traffic_agents_6}, a cautious agent waiting at the intersection even reverses to make room for an aggressive agent turning across its path. Similarly, in Fig.~\ref{fig:qualitative_traffic_agents_5} and Fig.~\ref{fig:qualitative_traffic_agents_1}, the cautious agents queue at the intersection and trail far behind the aggressive agents, which are the first to clear it.

\begin{table*}[ht]
\centering
\resizebox{\textwidth}{!}{
\begin{tabular}{l|ccccccc|ccccccc|ccccccc|ccccccc}
\toprule
\multirow{2}{*}{Method}
& \multicolumn{7}{c|}{IDM}
& \multicolumn{7}{c|}{SMART}
& \multicolumn{7}{c|}{PPO}
& \multicolumn{7}{c}{Expert} \\
 & AF & Off & Goal & Cmf & Aln & Ctr & Score & AF & Off & Goal & Cmf & Aln & Ctr & Score & AF & Off & Goal & Cmf & Aln & Ctr & Score & AF & Off & Goal & Cmf & Aln & Ctr & Score \\
\midrule
\multicolumn{29}{c}{\textit{Interactive1k}} \\
IDM     & 1.70 & 1.10 & 15.40 & 94.25 & 96.42 & 96.84 & 15.11 & 1.50 & 1.30 & 16.00 & 93.39 & 96.14 & 96.51 & 15.74 & 1.50 & 1.50 & 21.00 & 93.83 & 96.35 & 96.76 & 20.61 & 3.60 & 1.20 & 12.40 & 93.18 & 96.23 & 96.31 & 12.17 \\
PDM     & 1.70 & 0.70 & 24.80 & 90.62 & 95.73 & 95.95 & 24.11 & 1.00 & 0.70 & 31.20 & 88.69 & 95.26 & 94.66 & 30.44 & 0.90 & 1.10 & 30.60 & 84.06 & 95.47 & 94.35 & 29.23 & 3.10 & 0.90 & 21.20 & 89.31 & 95.40 & 94.92 & 20.68 \\
PPO     & 4.00 & 0.40 & 80.20 & 27.87 & 69.43 & 63.43 & 47.79 & 4.60 & 0.40 & 81.50 & 28.60 & 71.35 & 62.92 & 49.62 & 0.50 & 0.20 & 96.90 & 25.17 & 77.41 & 62.64 & 61.05 & 21.20 & 0.80 & 63.70 & 29.20 & 71.75 & 63.74 & 39.73 \\
SMART   & 14.30 & 5.00 & 34.00 & 48.20 & 91.02 & 82.86 & 28.13 & 4.30 & 6.80 & 44.70 & 47.89 & 90.28 & 82.12 & 36.76 & 3.50 & 6.10 & 41.20 & 48.00 & 90.61 & 82.19 & 33.84 & 8.70 & 5.30 & 46.40 & 47.90 & 90.40 & 82.16 & 38.29 \\
PDM+PPO & 0.80 & 0.10 & 76.70 & 45.72 & 84.75 & 81.47 & 59.17 & 2.30 & 0.20 & 84.00 & 44.06 & 85.28 & 80.89 & 64.89 & 0.30 & 0.00 & 95.30 & 47.56 & 87.90 & 83.02 & 75.25 & 11.50 & 0.20 & 70.20 & 44.60 & 84.64 & 79.88 & 54.27 \\
\midrule
\multicolumn{29}{c}{\textit{Random1k}} \\
IDM     & 1.60 & 2.90 & 57.40 & 93.84 & 94.72 & 93.34 & 55.40 & 1.80 & 2.80 & 54.80 & 93.28 & 94.78 & 93.35 & 52.94 & 1.40 & 2.80 & 61.30 & 93.53 & 94.73 & 93.35 & 59.11 & 4.60 & 3.00 & 52.60 & 93.36 & 94.65 & 93.08 & 50.80 \\
PDM     & 1.20 & 1.30 & 67.70 & 89.27 & 93.94 & 92.48 & 63.91 & 0.40 & 1.40 & 64.60 & 88.75 & 94.05 & 92.10 & 61.08 & 0.20 & 1.60 & 71.60 & 86.58 & 93.99 & 91.85 & 67.10 & 4.10 & 1.40 & 60.10 & 89.22 & 93.90 & 92.08 & 56.87 \\
PPO     & 1.40 & 0.20 & 92.20 & 26.48 & 79.42 & 62.97 & 59.86 & 1.70 & 0.40 & 89.00 & 26.84 & 79.24 & 62.84 & 57.48 & 0.30 & 0.00 & 98.00 & 24.40 & 84.25 & 63.30 & 65.17 & 9.50 & 0.40 & 82.00 & 27.22 & 79.56 & 63.55 & 54.06 \\
SMART   & 3.90 & 6.80 & 58.10 & 48.74 & 88.60 & 78.73 & 48.13 & 2.00 & 6.60 & 56.20 & 48.62 & 88.80 & 78.41 & 46.55 & 1.90 & 6.90 & 58.50 & 48.70 & 88.71 & 78.32 & 48.46 & 5.00 & 6.00 & 60.40 & 48.60 & 88.53 & 78.42 & 50.07 \\
PDM+PPO & 0.70 & 0.10 & 92.50 & 38.12 & 86.42 & 80.17 & 70.03 & 0.70 & 0.00 & 90.50 & 37.49 & 85.70 & 79.75 & 68.33 & 0.20 & 0.00 & 97.60 & 40.25 & 89.15 & 82.76 & 76.07 & 6.70 & 0.00 & 86.70 & 37.82 & 86.17 & 79.74 & 65.94 \\
\bottomrule
\end{tabular}
}
\caption{Planner performance against traffic agents on Interactive1k and Random1k. Each row is an ego planner (IDM, PDM, PPO, SMART, and our hybrid PDM+PPO), each column block a traffic agent controlling the surrounding agents: IDM, SMART, PPO, and Expert. We report at-fault collisions (AF), off-road rate (Off), goal-reaching (Goal), comfort (Cmf), alignment (Aln), centerline adherence (Ctr), and the aggregated driving score.}
\label{tab:planner_vs_traffic_reactive}
\end{table*}

\begin{table*}[ht]
\centering
\resizebox{\textwidth}{!}{
\begin{tabular}{l|ccccccc|ccccccc|ccccccc|ccccccc}
\toprule
\multirow{2}{*}{Method}
& \multicolumn{7}{c|}{Aggr.}
& \multicolumn{7}{c|}{Norm.}
& \multicolumn{7}{c|}{Caut.}
& \multicolumn{7}{c}{Mix} \\
 & AF & Off & Goal & Cmf & Aln & Ctr & Score & AF & Off & Goal & Cmf & Aln & Ctr & Score & AF & Off & Goal & Cmf & Aln & Ctr & Score & AF & Off & Goal & Cmf & Aln & Ctr & Score \\
\midrule
\multicolumn{29}{c}{\textit{Interactive1k}} \\
IDM     & 6.90 & 1.30 & 11.40 & 91.73 & 96.24 & 95.77 & 11.22 & 7.10 & 1.60 & 8.10 & 91.42 & 96.38 & 95.89 & 7.95 & 11.10 & 1.00 & 5.50 & 90.46 & 96.47 & 95.92 & 5.36 & 8.60 & 1.20 & 7.90 & 91.11 & 96.35 & 95.99 & 7.76 \\
PDM     & 3.40 & 4.60 & 21.30 & 84.97 & 95.26 & 93.91 & 20.76 & 4.40 & 3.30 & 14.40 & 84.69 & 95.30 & 94.17 & 14.02 & 6.90 & 3.30 & 8.70 & 83.60 & 95.31 & 94.02 & 8.42 & 5.80 & 5.10 & 13.10 & 84.42 & 95.36 & 94.06 & 12.69 \\
PPO     & 3.60 & 1.00 & 79.30 & 27.68 & 71.36 & 63.06 & 47.27 & 5.00 & 1.40 & 73.70 & 27.92 & 69.64 & 62.94 & 43.21 & 5.10 & 2.30 & 69.60 & 27.67 & 68.28 & 63.09 & 40.23 & 5.00 & 1.30 & 73.30 & 27.83 & 70.25 & 63.12 & 43.21 \\
SMART   & 17.90 & 4.30 & 28.70 & 48.58 & 91.39 & 83.18 & 23.68 & 22.90 & 3.50 & 22.80 & 48.64 & 91.21 & 83.06 & 18.77 & 35.70 & 3.10 & 12.00 & 48.92 & 91.79 & 83.37 & 9.96 & 26.70 & 3.80 & 18.10 & 48.73 & 91.39 & 83.29 & 14.92 \\
PDM+PPO & 1.90 & 1.60 & 73.90 & 41.88 & 80.21 & 75.70 & 53.83 & 2.80 & 2.20 & 67.10 & 39.52 & 78.42 & 74.04 & 47.33 & 3.60 & 3.90 & 60.10 & 34.56 & 75.63 & 71.70 & 39.70 & 2.60 & 2.50 & 65.00 & 37.69 & 78.55 & 73.49 & 45.34 \\
\midrule
\multicolumn{29}{c}{\textit{Random1k}} \\
IDM     & 3.20 & 2.80 & 51.40 & 92.57 & 94.63 & 93.23 & 49.50 & 4.50 & 2.80 & 48.50 & 92.15 & 94.63 & 93.20 & 46.64 & 7.00 & 2.60 & 44.00 & 91.63 & 94.76 & 93.34 & 42.24 & 4.50 & 2.70 & 47.80 & 92.03 & 94.76 & 93.34 & 45.98 \\
PDM     & 1.90 & 5.10 & 58.90 & 86.76 & 93.53 & 91.59 & 55.40 & 2.90 & 4.60 & 57.00 & 86.70 & 93.67 & 91.73 & 53.58 & 4.10 & 4.50 & 50.90 & 85.91 & 93.45 & 91.73 & 47.66 & 3.00 & 5.00 & 54.90 & 86.51 & 93.54 & 91.72 & 51.60 \\
PPO     & 0.90 & 0.30 & 91.10 & 25.73 & 79.77 & 62.18 & 58.47 & 1.70 & 0.70 & 86.80 & 25.86 & 78.21 & 62.29 & 55.02 & 3.30 & 1.20 & 83.80 & 26.19 & 76.79 & 62.07 & 52.45 & 2.10 & 1.10 & 86.80 & 25.85 & 78.38 & 62.01 & 55.21 \\
SMART   & 9.70 & 6.10 & 50.40 & 48.81 & 88.93 & 78.47 & 41.65 & 11.60 & 5.40 & 47.20 & 48.81 & 88.83 & 78.56 & 38.93 & 19.30 & 5.50 & 39.90 & 48.88 & 89.07 & 78.87 & 32.88 & 12.00 & 6.00 & 44.30 & 48.81 & 88.96 & 78.50 & 36.60 \\
PDM+PPO & 0.80 & 0.90 & 89.10 & 35.33 & 83.76 & 72.47 & 63.62 & 1.20 & 1.10 & 87.30 & 34.35 & 83.11 & 71.87 & 61.70 & 1.70 & 2.10 & 81.40 & 31.80 & 81.00 & 70.73 & 55.88 & 0.90 & 1.60 & 86.60 & 33.25 & 82.26 & 71.64 & 61.00 \\
\bottomrule
\end{tabular}
}
\caption{Planner performance against behavior-conditioned traffic agents on Interactive1k and Random1k. The four traffic policies share the same backbone but are conditioned on distinct driving styles: aggressive (Aggr.), normal (Norm.), and cautious (Caut.) behaviors, and a mixture (Mix) where each surrounding agent is independently assigned one of the three styles. We report at-fault collisions (AF), off-road rate (Off), goal-reaching (Goal), comfort (Cmf), alignment (Aln), centerline adherence (Ctr), and the aggregated driving score.}
\label{tab:planner_vs_traffic_behavior}
\end{table*}

\begin{table}[!ht]
\centering
\scriptsize
\setlength{\tabcolsep}{3pt}
\renewcommand{\arraystretch}{1.15}
\resizebox{1\textwidth}{!}{
\begin{tabular}{ll|ccccccccc|c|cc}
\toprule
\multirow{3}{*}{Split} & \multirow{3}{*}{Policy}
& \multicolumn{4}{c}{Kinematic}
& \multicolumn{3}{c}{Interactive}
& \multicolumn{2}{c|}{Map-based}
& \multirow{3}{*}{\makecell{Composite\\Metric$\uparrow$}}
& \multirow{3}{*}{ADE$\downarrow$}
& \multirow{3}{*}{minADE$\downarrow$} \\
\cmidrule(lr){3-6} \cmidrule(lr){7-9} \cmidrule(lr){10-11}
& & \makecell{Linear\\Speed$\uparrow$}
  & \makecell{Linear\\Accel.$\uparrow$}
  & \makecell{Ang.\\Speed$\uparrow$}
  & \makecell{Ang.\\Accel.$\uparrow$}
  & \makecell{Dist. to\\Obj.$\uparrow$}
  & \makecell{Collision$\uparrow$}
  & \makecell{TTC$\uparrow$}
  & \makecell{Dist. to\\Road Edge$\uparrow$}
  & \makecell{Offroad$\uparrow$}
  & & & \\
\midrule
\multirow{4}{*}{Interactive1k}
& $\pi_\theta^{\text{mix}}$
  & 0.060 & 0.143 & 0.416 & 0.540 & 0.332 & 0.952 & 0.791 & 0.435 & 0.918 & 0.681 & 8.11 & 8.11 \\
& \color{red}$\pi_\theta^{\text{aggr}}$
  & 0.131 & 0.197 & 0.445 & 0.545 & 0.362 & 0.901 & 0.778 & 0.500 & 0.889 & 0.677 & 5.39 & 5.39 \\
& \color{purple}$\pi_\theta^{\text{norm}}$
  & 0.099 & 0.168 & 0.419 & 0.540 & 0.345 & 0.975 & 0.805 & 0.450 & 0.920 & 0.695 & 5.96 & 5.96 \\
& \color{green}$\pi_\theta^{\text{caut}}$
  & 0.020 & 0.079 & 0.391 & 0.532 & 0.298 & 0.986 & 0.820 & 0.372 & 0.943 & 0.682 & 14.13 & 14.13 \\
\midrule
\multirow{4}{*}{Random1k}
& $\pi_\theta^{\text{mix}}$
  & 0.104 & 0.188 & 0.420 & 0.591 & 0.223 & 0.968 & 0.769 & 0.499 & 0.921 & 0.686 & 6.80 & 6.80 \\
& \color{red}$\pi_\theta^{\text{aggr}}$
  & 0.149 & 0.221 & 0.450 & 0.597 & 0.256 & 0.924 & 0.765 & 0.547 & 0.912 & 0.686 & 5.10 & 5.10 \\
& \color{purple}$\pi_\theta^{\text{norm}}$
  & 0.131 & 0.206 & 0.426 & 0.592 & 0.240 & 0.986 & 0.777 & 0.511 & 0.924 & 0.698 & 5.30 & 5.30 \\
& \color{green}$\pi_\theta^{\text{caut}}$
  & 0.066 & 0.142 & 0.392 & 0.583 & 0.193 & 0.992 & 0.777 & 0.457 & 0.937 & 0.684 & 10.92 & 10.92 \\
\bottomrule
\end{tabular}}
\caption{\textbf{Per-subscore WOSAC 2024 for the four conditioned policies on Interactive1k and Random1k.}}
\label{tab:realism_subscores}
\end{table}

\clearpage
\section*{NeurIPS Paper Checklist}

\begin{enumerate}

\item {\bf Claims}
    \item[] Question: Do the main claims made in the abstract and introduction accurately reflect the paper's contributions and scope?
    \item[] Answer: \answerYes{} 
    \item[] Justification: All claims and contributions stated in the abstract and introduction are described in detail in the main paper and supported by our empirical results.

\item {\bf Limitations}
    \item[] Question: Does the paper discuss the limitations of the work performed by the authors?
    \item[] Answer: \answerYes{} 
    \item[] Justification: We discuss the limitations of BehaviorBench in the conclusion paragraph, including the lack of traffic lights and stop signs.

\item {\bf Theory assumptions and proofs}
    \item[] Question: For each theoretical result, does the paper provide the full set of assumptions and a complete (and correct) proof?
    \item[] Answer: \answerNA{} 
    \item[] Justification: The paper does not include any theoretical results and therefore contains no proofs.

    \item {\bf Experimental result reproducibility}
    \item[] Question: Does the paper fully disclose all the information needed to reproduce the main experimental results of the paper to the extent that it affects the main claims and/or conclusions of the paper (regardless of whether the code and data are provided or not)?
    \item[] Answer: \answerYes{} 
    \item[] Justification: The main paper describes the benchmark splits, traffic agents, evaluation metrics, and our hybrid planner. Hyperparameters and training details are provided in the Appendix. We also provide the code to reproduce the results.

\item {\bf Open access to data and code}
    \item[] Question: Does the paper provide open access to the data and code, with sufficient instructions to faithfully reproduce the main experimental results, as described in supplemental material?
    \item[] Answer: \answerYes{} 
    \item[] Justification: We release the code for generating our scenario splits and the full evaluation framework, enabling reproduction of all reported results.

\item {\bf Experimental setting/details}
    \item[] Question: Does the paper specify all the training and test details (e.g., data splits, hyperparameters, how they were chosen, type of optimizer) necessary to understand the results?
    \item[] Answer: \answerYes{} 
    \item[] Justification: All training and evaluation details are provided in the main paper and appendix.

\item {\bf Experiment statistical significance}
    \item[] Question: Does the paper report error bars suitably and correctly defined or other appropriate information about the statistical significance of the experiments?
    \item[] Answer: \answerYes{} 
    \item[] Justification: For PPO and PDM+PPO we report mean±std over three training runs with different seeds (Tab. ~\ref{tab:planner_score}).

\item {\bf Experiments compute resources}
    \item[] Question: For each experiment, does the paper provide sufficient information on the computer resources (type of compute workers, memory, time of execution) needed to reproduce the experiments?
    \item[] Answer: \answerYes{} 
    \item[] Justification: We report the GPU type, training time, and memory requirements in the Appendix.
    
\item {\bf Code of ethics}
    \item[] Question: Does the research conducted in the paper conform, in every respect, with the NeurIPS Code of Ethics \url{https://neurips.cc/public/EthicsGuidelines}?
    \item[] Answer: \answerYes{} 
    \item[] Justification: Our work uses publicly available driving datasets and simulators, involves no human subjects or personal data, and conforms with the NeurIPS Code of Ethics in every respect.

\item {\bf Broader impacts}
    \item[] Question: Does the paper discuss both potential positive societal impacts and negative societal impacts of the work performed?
    \item[] Answer: \answerNA{} 
    \item[] Justification: The paper introduces an evaluation framework and benchmark for autonomous driving planners. As a methodological contribution that does not deploy a system, it has no direct societal impact beyond that of the underlying research area.
    
\item {\bf Safeguards}
    \item[] Question: Does the paper describe safeguards that have been put in place for responsible release of data or models that have a high risk for misuse (e.g., pre-trained language models, image generators, or scraped datasets)?
    \item[] Answer: \answerNA{} 
    \item[] Justification: The released artifacts are evaluation code and benchmark splits, which do not pose a high risk of misuse.

\item {\bf Licenses for existing assets}
    \item[] Question: Are the creators or original owners of assets (e.g., code, data, models), used in the paper, properly credited and are the license and terms of use explicitly mentioned and properly respected?
    \item[] Answer: \answerYes{} 
    \item[] Justification: All datasets, simulators, and planner implementations used in this work are cited at first use, and we comply with their respective licenses and terms of use.

\item {\bf New assets}
    \item[] Question: Are new assets introduced in the paper well documented and is the documentation provided alongside the assets?
    \item[] Answer: \answerYes{} 
    \item[] Justification: The released code includes documentation covering the scenario splits, evaluation framework, and usage instructions to reproduce our results.

\item {\bf Crowdsourcing and research with human subjects}
    \item[] Question: For crowdsourcing experiments and research with human subjects, does the paper include the full text of instructions given to participants and screenshots, if applicable, as well as details about compensation (if any)? 
    \item[] Answer: \answerNA{} 
    \item[] Justification:  The paper does not involve crowdsourcing or research with human subjects.

\item {\bf Institutional review board (IRB) approvals or equivalent for research with human subjects}
    \item[] Question: Does the paper describe potential risks incurred by study participants, whether such risks were disclosed to the subjects, and whether Institutional Review Board (IRB) approvals (or an equivalent approval/review based on the requirements of your country or institution) were obtained?
    \item[] Answer: \answerNA{} 
    \item[] Justification: The paper does not involve research with human subjects and therefore does not require IRB approval.

\item {\bf Declaration of LLM usage}
    \item[] Question: Does the paper describe the usage of LLMs if it is an important, original, or non-standard component of the core methods in this research? Note that if the LLM is used only for writing, editing, or formatting purposes and does \emph{not} impact the core methodology, scientific rigor, or originality of the research, declaration is not required.
    \item[] Answer: \answerNA{} 
    \item[] Justification: LLMs were not used as part of the core methodology.

\end{enumerate}

\end{document}